%% file: aaai24.tex
\documentclass[letterpaper]{article} 
\usepackage{aaai24}  
\usepackage{times}  
\usepackage{helvet}  
\usepackage{courier}  
\usepackage[hyphens]{url}  
\usepackage{graphicx} 
\usepackage{natbib}  
\usepackage{caption} 
\usepackage{algorithm}
\usepackage{algorithmic}

\usepackage{newfloat}
\usepackage{listings}
\DeclareCaptionStyle{ruled}{labelfont=normalfont,labelsep=colon,strut=off} 
\floatstyle{ruled}
\newfloat{listing}{tb}{lst}{}
\floatname{listing}{Listing}
\input{math_commands.tex}

\usepackage{subcaption}
\usepackage{amsmath}
\usepackage{amsfonts}
\usepackage{amssymb}
\usepackage{mathtools}
\usepackage{amsthm}
\usepackage{multirow}
\usepackage{nicefrac}
\usepackage{url}
\usepackage{xcolor}
\usepackage{comment}

\newtheorem{theorem}{Theorem}
\newtheorem{proposition}[theorem]{Proposition}

\newtheorem{definition}[theorem]{Definition}

\title{Parameterized Projected Bellman Operator}
\author{
    Théo Vincent\textsuperscript{\rm 1}\textsuperscript{\rm 2}\thanks{Correspondence to: theo.vincent@dfki.de},
    Alberto Maria Metelli\textsuperscript{\rm 3},
    Boris Belousov\textsuperscript{\rm 1},\\
    Jan Peters\textsuperscript{\rm 1}\textsuperscript{\rm 2}\textsuperscript{\rm 4}\textsuperscript{\rm 5},
    Marcello Restelli\textsuperscript{\rm 3},
    Carlo D'Eramo\textsuperscript{\rm 2}\textsuperscript{\rm 4}\textsuperscript{\rm 6}
}
\affiliations{
    \textsuperscript{\rm 1}German Research Center for AI (DFKI), Germany \\
    \textsuperscript{\rm 2}Department of Computer Science, TU Darmstadt, Darmstadt, Germany \\
    \textsuperscript{\rm 3}Department of Electronics, Computer Science, and Bioengineering, Politecnico di Milano, Milano, Italy \\
    \textsuperscript{\rm 4}Hessian.ai, Germany \\
    \textsuperscript{\rm 5}Centre for Cognitive Science, TU Darmstadt, Darmstadt, Germany \\
    \textsuperscript{\rm 6}Center for Artificial Intelligence and Data Science, University of Würzburg, Würzburg, Germany

}

\begin{document}

\maketitle

\begin{abstract}
Approximate value iteration~(AVI) is a family of algorithms for reinforcement learning~(RL) that aims to obtain an approximation of the optimal value function. Generally, AVI algorithms implement an iterated procedure where each step consists of (i) an application of the Bellman operator and (ii) a projection step into a considered function space. Notoriously, the Bellman operator leverages transition samples, which strongly determine its behavior, as uninformative samples can result in negligible updates or long detours, whose detrimental effects are further exacerbated by the computationally intensive projection step. To address these issues, we propose a novel alternative approach based on learning an approximate version of the Bellman operator rather than estimating it through samples as in AVI approaches. This way, we are able to (i) generalize across transition samples and (ii) avoid the computationally intensive projection step. For this reason, we call our novel operator \textit{projected Bellman operator}~(PBO). We formulate an optimization problem to learn PBO for generic sequential decision-making problems, and we theoretically analyze its properties in two representative classes of RL problems. Furthermore, we theoretically study our approach under the lens of AVI and devise algorithmic implementations to learn PBO in offline and online settings by leveraging neural network parameterizations. Finally, we empirically showcase the benefits of PBO w.r.t. the regular Bellman operator on several RL problems.
\end{abstract}

\section{Introduction}
Value-based reinforcement learning~(RL) is a popular class of algorithms for solving sequential decision-making problems with unknown dynamics~\citep{sutton2018reinforcement}. For a given problem, value-based algorithms aim at obtaining the most accurate estimate of the expected return from each state, i.e., a value function. For instance, the well-known value-iteration algorithm computes value functions by iterated applications of the Bellman operator~\citep{bellman1966dynamic}, of which the true value function is the fixed point. Although the Bellman operator can be applied in an exact way in dynamic programming, it is typically estimated from samples \textit{at each application} when dealing with problems with unknown models, i.e., empirical Bellman operator~\citep{watkins1989learning,bertsekas2019reinforcement}. Intuitively, the dependence of the empirical version of value iteration on the samples has an impact on the efficiency of the algorithms and on the quality of the obtained estimated value function, which becomes accentuated when solving continuous problems that require function approximation, e.g., approximate value iteration~(AVI)~\citep{munos2005error,munos2008finite}. Moreover, in AVI approaches, costly function approximation steps are needed to project the output of the Bellman operator back to the considered value function space.

\begin{figure}[b]
    \centering
    \includegraphics[width=0.425\columnwidth]{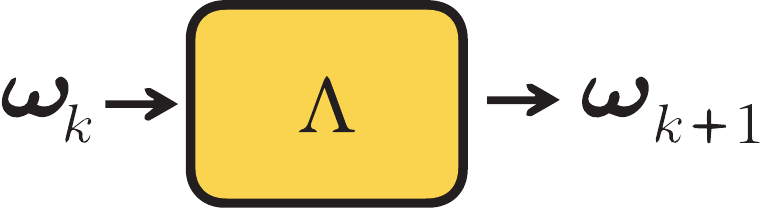}
    \hspace{.15cm}
    \includegraphics[width=0.53\columnwidth]{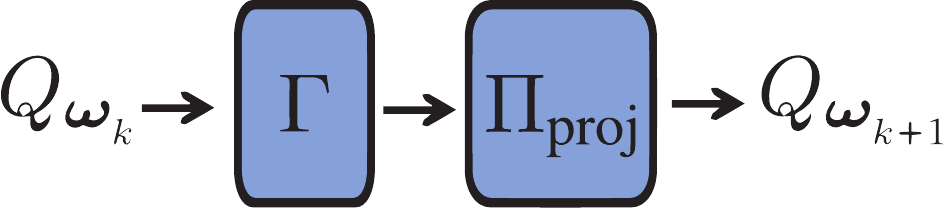}
    \caption{PBO $\Lambda$ (left) operates on value function parameters as opposed to AVI (right), that uses the empirical Bellman operator $\Gamma$ followed by the projection operator $\Pi_{\text{proj}}$.}
    \label{F:bo_vs_pbo}
\end{figure}

\begin{figure*}
    \centering
    \begin{subfigure}{0.49\textwidth}
        \centering
        \includegraphics[width=.8\columnwidth]{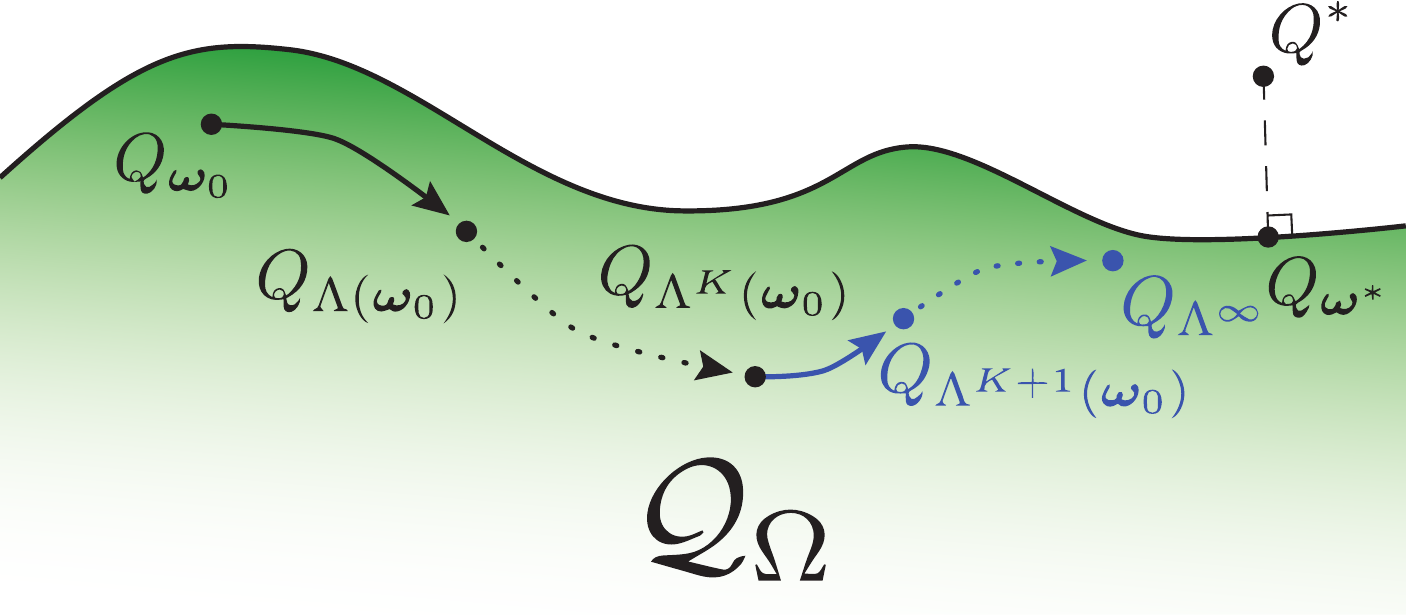}
        \caption{Projected Bellman Operator~(PBO) (ours)}\label{F:pbo}
    \end{subfigure}
    \begin{subfigure}{0.49\textwidth}
        \centering
        \includegraphics[width=.8\columnwidth]{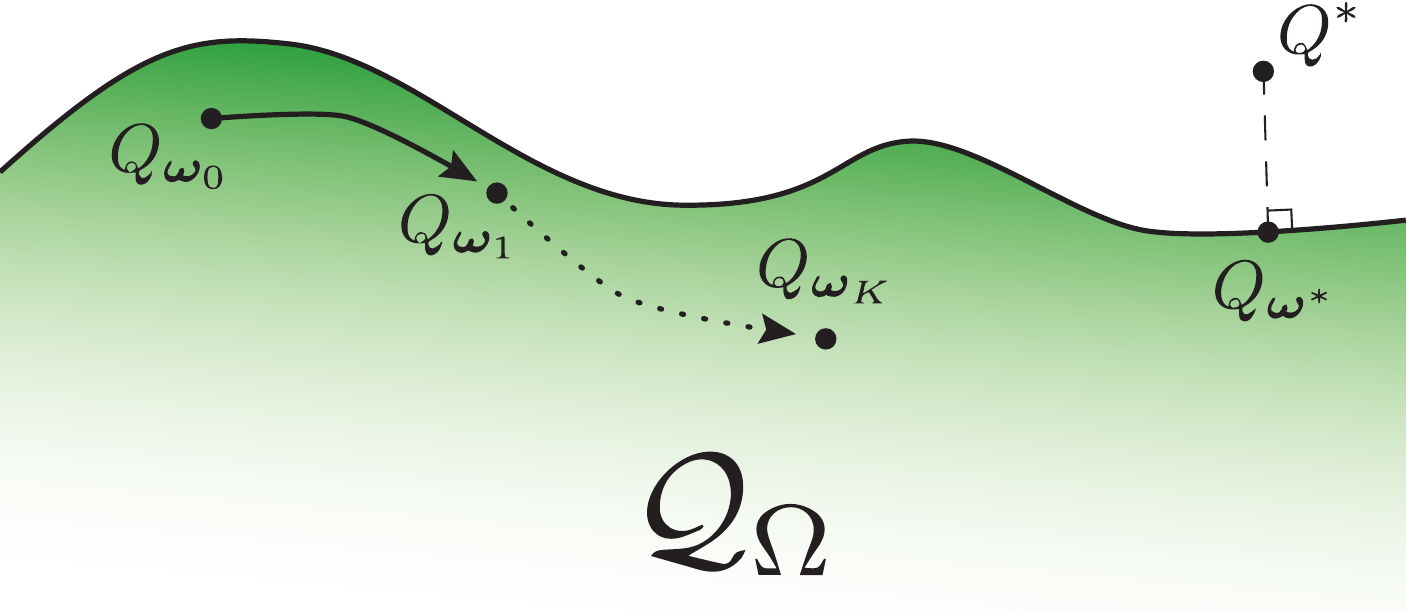}
        \caption{Approximate Value Iteration~(AVI)}\label{F:avi}
    \end{subfigure}
    \caption{Behavior of our PBO and AVI in the parametric space of value functions $\mathcal{Q}$. Here $Q^*$, $Q_{\vomega^*}$, and $Q_{\Lambda^\infty}$, are respectively the optimal value function, its projection on the parametric space, and the fixed point of PBO. Contrary to the regular Bellman operator, PBO can be applied for an arbitrary number of steps (blue lines) without requiring additional samples.}
    \label{F:pbo_avi}
\end{figure*}

In this paper, we introduce the \textit{projected Bellman operator}~(PBO), which consists of a function $\Lambda: \Omega \to \Omega$ defined on the parameters $\vomega \in \Omega$ of the value function approximator $Q_\vomega$. Contrary to the standard (empirical) Bellman operator $\Gamma$, which acts on the value functions $Q_{k}$ to compute targets that are then projected to obtain $Q_{k+1}$, our PBO $\Lambda$ acts on the parameters of the value function to directly compute updated parameters $\vomega_{k+1} = \Lambda(\vomega_k)$ (Figure~\ref{F:bo_vs_pbo}). The advantages of our approach are twofold: (i) PBO is applicable for an arbitrary number of iterations without using further samples, and (ii) the output of PBO always belongs to the considered value function space as visualized in Figure~\ref{F:pbo_avi}, thus avoiding the costly projection step which is required when using the Bellman operator coupled with function approximation. We show how to estimate PBO from transition samples by leveraging a parametric approximation which we call \emph{parameterized} PBO, and we devise two algorithms for offline and online RL to learn it.
Starting from initial parameters $\vomega_0$, AVI approaches obtain consecutive approximations of the value function $Q_{\vomega_k}$ by applying the Bellman operator iteratively over samples (Figure~\ref{F:avi}). Instead, we make use of the samples to learn the PBO only. Then, starting from initial parameters $\vomega_0$, PBO can produce a chain of value function parameters of \textit{arbitrary} length (as shown with the blue lines in Figure~\ref{F:pbo}) without requiring further samples. This means an accurate approximation of PBO can compute optimal value function parameters starting from any initial parameters in the chosen space without requiring additional samples.

\section{Related Work}
Several papers in the literature proposed variants of the standard Bellman operator to induce certain behaviors. We review these approaches, noting that they all act on the space of action-value functions, thus needing a costly projection step onto the considered function space. Conversely, to the best of our knowledge, our work is the first attempt to obtain an alternative Bellman operator that avoids the projection step by directly acting on the parameters of value functions.

\textbf{Bellman operator variations.}~Variants of the Bellman operator are widely studied for entropy-regularized MDPs~\citep{neu2017unified,geist2019theory,belousov2019entropic}. The \textit{softmax}~\citep{haarnoja2017reinforcement,song2019revisiting}, \textit{mellowmax}~\citep{asadi2017alternative}, and \textit{optimistic}~\citep{tosatto2019exploration} operators are all examples of variants of the Bellman operator to obtain maximum-entropy exploratory policies. Besides favoring exploration, other approaches address the limitations of the standard Bellman operator. For instance, the \textit{consistent} Bellman operator~\citep{bellemare2016increasing} is a modified operator that addresses the problem of inconsistency of the optimal action-value functions for suboptimal actions. The \textit{distributional} Bellman operator~\citep{bellemare2017distributional} enables operating on the whole return distribution instead of its expectation, i.e., the value function~\citep{bdr2022}. Furthermore, the \textit{logistic} Bellman operator uses a logistic loss to solve a convex linear programming problem to find optimal value functions~\citep{bas2021logistic}. Finally, the \textit{Bayesian} Bellman operator is employed to infer a posterior over Bellman operators centered on the true one~\citep{fellows2021bayesian}. Note that our PBO can be seamlessly applied to any of these variants of the standard Bellman operator. Finally, we recognize that learning an approximation of the Bellman operator shares some similarities with learning a reward-transition model in reinforcement learning. However, we point out that our approach is profoundly different, as we map action-value parameters to other action-value parameters, in contrast to model-based reinforcement learning, which maps states and actions to rewards and next states. 

\textbf{Operator learning.}~Literature in operator learning is mostly focused on supervised learning, with methods for learning operators over vector spaces~\citep{micchelli2005learning} and parametric approaches for learning non-linear operators~\citep{chen1995universal}, with a resurgence of recent contributions in deep learning. For example,~\citet{kovachki2021neural,kovachki2023neural} learn mappings between infinite function spaces with deep neural networks, or~\citet{kissas2022learning} apply an attention mechanism to learn correlations in the target function for efficient operator learning. We note that our work on the learning of the Bellman operator in reinforcement learning is orthogonal to methods for operator learning in supervised learning, and could potentially benefit from advanced techniques in the literature. While our work focuses on learning the Bellman operator in the parameter space, \citet{Tamar2016} approximate the Bellman operator in the function space.

\textbf{HyperNetworks.}~Our approach shares similarities with HyperNetworks \citep{HyperNetworks,sarafian21a,beck23a}. In the classification proposed in \citet{chauhan2023brief}, PBO distinguishes itself by taking as input the parameters generated at the previous forward pass. In this sense, PBO can be seen as a recurrent variant of HyperNetworks that generates a series of parameters representing the consecutive Bellman iterations.

\section{Preliminaries}
\label{S:preliminaries}
We consider discounted Markov decision processes~(MDPs) defined as $\mathcal{M} = \langle \mathcal{S}$, $\mathcal{A}$, $\mathcal{P}$, $\mathcal{R}$, $\gamma \rangle$, where $\mathcal{S}$ is a measurable state space, $\mathcal{A}$ is a finite, measurable action space, $\mathcal{P}:\mathcal{S} \times \mathcal{A} \to \Delta(\mathcal{S})$\footnote{$\Delta(\mathcal{X})$ denotes the set of probability measures over the set $\mathcal{X}$.} is the transition kernel of the dynamics of the system, $\mathcal{R}:\mathcal{S} \times \mathcal{A} \to \mathbb{R}$ is a reward function, and $\gamma \in [0,1)$ is a discount factor~\citep{puterman1990markov}. A policy $\pi: \mathcal{S} \to \mathcal{A}$ is a function mapping a state to an action, inducing a value function $V^{\pi}(s) \triangleq \mathbb{E}\left[\sum_{t=0}^{+\infty} \gamma^t \mathcal{R}(S_t,\pi(S_t)) | S_0=s\right]$ representing the expected cumulative discounted reward starting in state $s$ and following policy $\pi$ thereafter. Similarly, the action-value function $Q^\pi(s,a) \triangleq \mathbb{E}\left[\sum_{t=0}^{+\infty} \gamma^t \mathcal{R}(S_t, A_t) | S_0=s, A_0=a, A_t = \pi(S_t)\right]$ is the expected discounted cumulative reward executing action $a$ in state $s$, following policy $\pi$ thereafter. RL aims to find an optimal policy $\pi^*$ yielding the optimal value function $V^*(s) \triangleq \max_{\pi:\mathcal{S} \to \mathcal{A}} V^{\pi}(s)$ for every $ s \in \mathcal{S}$~\citep{puterman1990markov}.
The (optimal) Bellman operator $\Gamma^*$ is a fundamental tool in RL for obtaining optimal policies, defined as:
\begin{equation}\label{E:opt_bellman}\resizebox{.9\linewidth}{!}{$\displaystyle
    (\Gamma^*Q)(s,a) \triangleq \mathcal{R}(s,a) + \gamma \int_\mathcal{S} \mathcal{P}(\mathrm{d}s'|s,a)\max_{a' \in \mathcal{A}} Q(s',a'),$}
\end{equation}
for all $(s,a) \in \mathcal{S\times A}$.
It is well-known that Bellman operators are $\gamma$-contraction mappings in $L_{\infty}$-norm, such that their iterative application leads to the unique fixed point $\Gamma^* Q^*=Q^*$ in the limit~\citep{Bertsekas2015DynamicPA}. We consider using function approximation to represent value functions and denote $\Omega$ as the space of their parameters. Thus, we define $\mathcal{Q}_{\Omega} = \left\lbrace Q_{\vomega}: \mathcal{S\times A}\to\mathbb{R} \vert \vomega \in \Omega \right\rbrace$ as the set of value functions representable by parameters of $\Omega$.

\section{Projected Bellman Operator}\label{S:method}
The application of the Bellman operator in RL requires transition samples (Equation~\ref{E:opt_bellman}) and a costly projection step onto the used function space ~\citep{munos2003error,munos2005error,munos2008finite}. We are interested in obtaining an operator that overcomes these issues while emulating the behavior of the regular Bellman operator. Hence, we introduce the \textit{projected Bellman operator}~(PBO), which we define as follows.
\begin{definition}
Let $\mathcal{Q}_{\Omega} = \left\lbrace Q_{\vomega}: \mathcal{S\times A}\to\mathbb{R} \vert \vomega \in \Omega \right\rbrace$ be a function approximation space for the action-value function, induced by the parameter space $\Omega$. A projected Bellman operator~(PBO) is a function $\Lambda:\Omega \to \Omega$, such that
\begin{equation}\label{E:opt_pbo}\resizebox{.9\linewidth}{!}{$\displaystyle
    \Lambda \in \argmin_{\Lambda : \Omega \to \Omega} \mathbb{E}_{(s,a) \sim \rho, \vomega \sim \nu} \left(\Gamma^*Q_{\vomega}(s,a) - Q_{\Lambda(\vomega)}(s,a)\right)^2, $}
\end{equation}
for state-action and parameter distributions $\rho$ and $\nu$.
\end{definition}
Note that $\Gamma^*$ is the regular optimal Bellman operator on $\mathcal{Q}_{\Omega}$. Conversely, PBO is an operator $\Lambda$ acting on action-value function parameters $\vomega \in \Omega$. In other words, this definition states that a PBO is the mapping $\Omega \to \Omega$ that most closely emulates the behavior of the regular optimal Bellman operator $\Gamma^*$. Thus, by acting on the parameters $\vomega$ of action-value functions $Q_\vomega$, PBO can be applied for an arbitrary number of steps starting from any initial parameterizations $\vomega_0$, without using additional transition samples. Moreover, being a mapping between parameters $\vomega_k$ to parameters $\vomega_{k+1}$, PBO does not require the costly projection step needed by the regular Bellman operator.
\begin{algorithm}
    \caption{Projected FQI \& *Projected DQN*}\label{A:algorithm}
    \begin{algorithmic}[1]
        \STATE \textbf{Inputs: }\\
            - samples $\mathcal{D} = \lbrace\langle s_j, a_j, r_j, s'_j \rangle\rbrace_{j=1}^J$;\\- parameters $\mathcal{W} = \lbrace\vomega_l\rbrace_{l=1}^L$;
            \\- \#Bellman iterations $K$;
            \\- initial parameters $\vphi$ of parameterized PBO $\Lambda_\vphi$;
            \\- \#Epochs $E$.
        \FOR{$e \in \{1,\dots,E\}$}
            \STATE $\bar{\vphi} = \vphi$
            \FOR{some training steps}
                \STATE *Collect samples $\mathcal{D}'$ with policy given by $Q_{\Lambda_\vphi^K(\vomega)}$*
                \STATE *$\mathcal{D} \gets \mathcal{D} \cup \mathcal{D}'$*
                \STATE Gradient descent over parameters $\vphi$ minimizing loss~(\ref{E:pbo_loss}) or~(\ref{E:pbo_loss_plus}) using $\mathcal{D}$ *or a batch of $\mathcal{D}$* and $\mathcal{W}$.
            \ENDFOR
        \ENDFOR
        \STATE \textbf{Return:} Parameters $\vphi$ of parameterized PBO $\Lambda_\vphi$
    \end{algorithmic}
\end{algorithm}
\begin{figure*}
    \centering
    \begin{subfigure}{0.49\textwidth}
        \centering
        \includegraphics[width=.8\columnwidth]{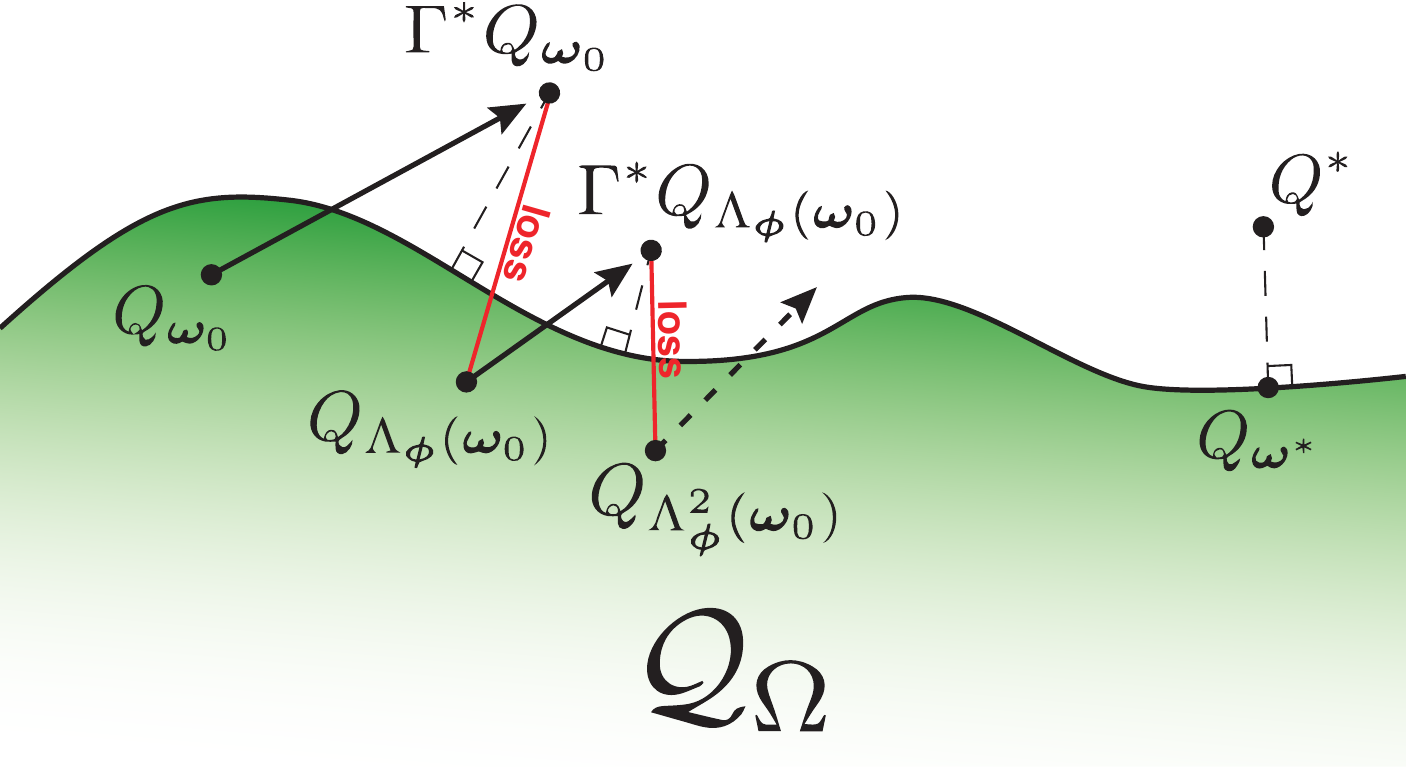}
        \caption{ProFQI (ours)}\label{F:alg_profqi}
    \end{subfigure}
    \begin{subfigure}{0.49\textwidth}
        \centering
        \includegraphics[width=.8\columnwidth]{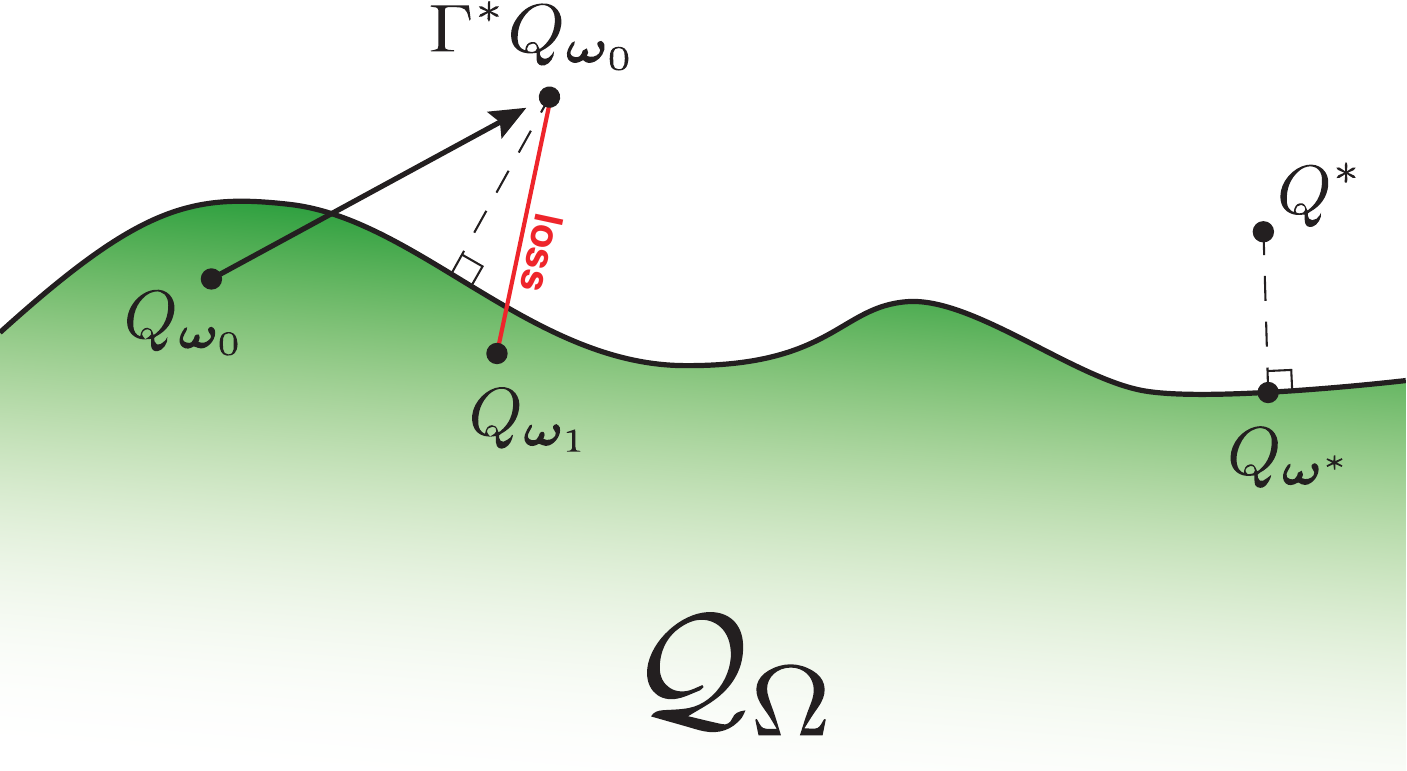}
        \caption{FQI}\label{F:alg_fqi}
    \end{subfigure}
    \caption{Behavior of ProFQI and FQI in the function space $\mathcal{Q}_\Omega$ for one iteration. The ability to apply PBO for an arbitrary number of times enables ProFQI to generate a sequence of action-value functions $Q_{\Lambda^k_\phi}(\vomega_0)$ that can be used to enrich the loss function to learn PBO (see red lines). On the contrary, one iteration of FQI corresponds to a single application of the Bellman operator followed by the projection step onto the function space.}
    \label{F:algorithms_training}
\end{figure*}
\subsection{Learning Projected Bellman Operators}
The PBO is unknown and has to be estimated from samples. We propose to approximate PBO with a \textit{parameterized PBO} $\Lambda_\vphi$ differentiable w.r.t. its parameters $\vphi \in \Phi$, enabling the use of gradient descent on a loss function.\footnote{For ease of presentation, we use PBO and parameterized PBO interchangeably whenever clear from the context.} We do so by formulating an empirical version of the problem~(\ref{E:opt_pbo}) that can be optimized via gradient descent by using a given dataset of parameters $\vomega \in \mathcal{W} \sim \nu$ and transitions $(s,a,r,s') \in \mathcal{D} \sim \rho$.
Crucially, we can leverage PBO to augment the dataset of parameters $\vomega\in\mathcal{W}$ with sequences generated by PBO at no cost of additional samples. The resulting loss is
\begin{equation}\label{E:pbo_loss}\resizebox{.9\linewidth}{!}{$\displaystyle
    \mathcal{L}_{\Lambda_\vphi}=\sum_{k = 1}^K \sum\limits_{\substack{(s,a)\in\mathcal{D}\\\vomega\in\mathcal{W}}} \left(\Gamma Q_{\Lambda^{k - 1}_\vphi(\vomega)}(s,a) - Q_{\Lambda^k_\vphi(\vomega)}(s,a) \right)^2,$}
\end{equation}
where $K$ is an arbitrary number of optimal Bellman operator iterations. Note that for $K=1$ we obtain the empirical version of the optimization problem~(\ref{E:opt_pbo}).
An additional idea is to add a term that corresponds to an infinite number of iterations, i.e., the fixed point,
\begin{equation}\label{E:pbo_loss_plus}\resizebox{.9\linewidth}{!}{$\displaystyle
    \mathcal{L}_{\Lambda^\infty_\vphi} = \mathcal{L}_{\Lambda_\vphi} + \sum_{(s,a) \in \mathcal{D}} \underbrace{\left( \Gamma Q_{\Lambda^\infty_\vphi}(s,a) - Q_{\Lambda^\infty_\vphi}(s,a) \right)^2}_{\text{Fixed point term}},$}
\end{equation}
where $\Lambda^\infty_\vphi$ is the fixed point of the parameterized PBO. Note that the addition of the fixed-point term is only possible for classes of parameterized PBOs where the fixed point can be computed, and the result is differentiable w.r.t. the parameters $\vphi$, as we describe in the following section.

We can now devise two algorithms to learn PBO in both offline and online RL. Our algorithms can be seen as variants of the offline algorithm fitted $Q$-iteration (FQI)~\citep{ernst2005fqi} and online algorithm deep $Q$-network (DQN)~\citep{mnih2015human}; thus, we call them \textit{projected fitted $Q$-iteration}~(ProFQI) and \textit{projected deep $Q$-network}~(ProDQN).

Algorithm~\ref{A:algorithm} compactly describes both ProFQI and ProDQN, highlighting the additional steps required for the online setting (i.e., ProDQN).
Both ProFQI and ProDQN are given initial randomly sampled datasets of transitions and parameters of PBO. As an online algorithm, ProDQN periodically adds new transitions to the dataset by executing a policy derived from the action-value function obtained by applying the current approximation of PBO for $K$ times. For stability reasons, we perform gradient descent steps only on the parameters $\vphi$ of $Q_{\Lambda_\vphi^k}$ in the loss (Equation~\ref{E:pbo_loss}), excluding the ones corresponding to the target $\Gamma Q_{\Lambda_\vphi^{k-1}}$, as commonly done in semi-gradient methods~\cite{sutton2018reinforcement}. Similar to~\citet{mnih2015human}, the target parameters are updated periodically after an arbitrary number of iterations. Soft updates, e.g., Polyak averaging~\cite{lillicrap2015continuous}, can also be used. As also illustrated by Figure~\ref{F:algorithms_training}, ProFQI and FQI behave substantially differently in the space of action-value functions $\mathcal{Q}_\Omega$. ProFQI aims to learn PBO to subsequently use it to generate a sequence $(Q_{\vomega_0},Q_{\Lambda_\phi(\vomega_0)},Q_{\Lambda^2_\phi(\vomega_0)},\dots)$ of action-value parameters starting from any parameters $\vomega_0$; on the contrary, FQI generates a sequence of action-value functions $(Q_{\vomega_0},Q_{\vomega_1},\dots)$ that need to be projected onto the function space at every iteration. Moreover, ProFQI can apply PBO multiple times to form a richer loss than the one for FQI, which can only consider one application of the Bellman operator at a time (see red lines in Figure~\ref{F:alg_profqi} and~\ref{F:alg_fqi}). 
\section{Analysis of Projected Bellman Operator}\label{S:analysis}
Besides the practical advantages of being independent of samples and not needing a projection step, PBO plays an interesting role when investigated theoretically in the AVI framework. In particular, the ability of PBO to iterate for an arbitrary number of times enables us to theoretically prove its benefit in terms of approximation error at a given timestep $K$ by leveraging the following established results in AVI.
\begin{theorem}\textit{(See Theorem 3.4 of \citet{farahmand2011regularization})}\label{Th:error_propagation} Let $K \in \sN^*$, $\rho, \nu$ two distribution probabilities over $\mathcal{S} \times \mathcal{A}$. For any sequence $(Q_k)_{k=0}^K \subset B \left(\mathcal{S} \times \mathcal{A}, R_{\gamma} \right)$ where $R_{\gamma}$ depends on reward function and discount factor, we have
\begin{align}
\| &Q^* - Q^{\pi_K} \|_{1, \rho} \leq C_{K, \gamma, R_{\gamma}} \label{E:error_prop} \\ + &\inf_{r \in [0, 1]} F(r; K, \rho, \gamma) \Biggl( \overbrace{\sum_{k=1}^{K} \alpha_k^{2r} \underbrace{\| \Gamma^*Q_{k - 1} - Q_k \|_{2, \nu}^{2}}_{FQI}}^{ProFQI} \Biggr)^{\frac{1}{2}},\notag
\end{align}
where $\alpha_k$ and $C_{K, \gamma, R_{\gamma}}$ do not depend on the sequence $(Q_k)_{k=0}^K$. $F(r; K, \rho, \gamma)$ relies on the concentrability coefficients of the greedy policies w.r.t. the value functions.
\end{theorem}
This theorem shows that the distance between the value function and the optimal value function (left-hand side of Equation~(\ref{E:error_prop})) is upper-bounded by a quantity proportional to the approximation errors $\|\Gamma^*Q_{k - 1} - Q_k \|_{2, \nu}^{2}$ at each iteration $k$. As highlighted in Equation~(\ref{E:error_prop}), at every iteration $k$, the loss of FQI contains only one term of the sum. Conversely, the loss of ProFQI contains the entire sum of approximation errors from iteration $k=1$ to $K$. This means that while FQI minimizes the approximation error for a single iteration at a time, ProFQI minimizes the approximation errors for multiple iterations, thus effectively reducing the upper bound on the approximation error at iteration $K$.

In the following, we empirically showcase this theoretical advantage and we further analyze the properties of PBO for two representative classes of RL problems, namely (i) finite MDPs and (ii) linear quadratic regulators~\citep{bradtke1992reinforcement,pang2021robust}. Proofs of the following results and additional analysis of PBO in low-rank MDPs~\citep{agarwal2020flambe,sekhari2021agnostic} can be found in the appendix.

\begin{figure*}
    \centering
    \begin{subfigure}{0.33\textwidth}
        \begin{center}
        \includegraphics[scale=.35]{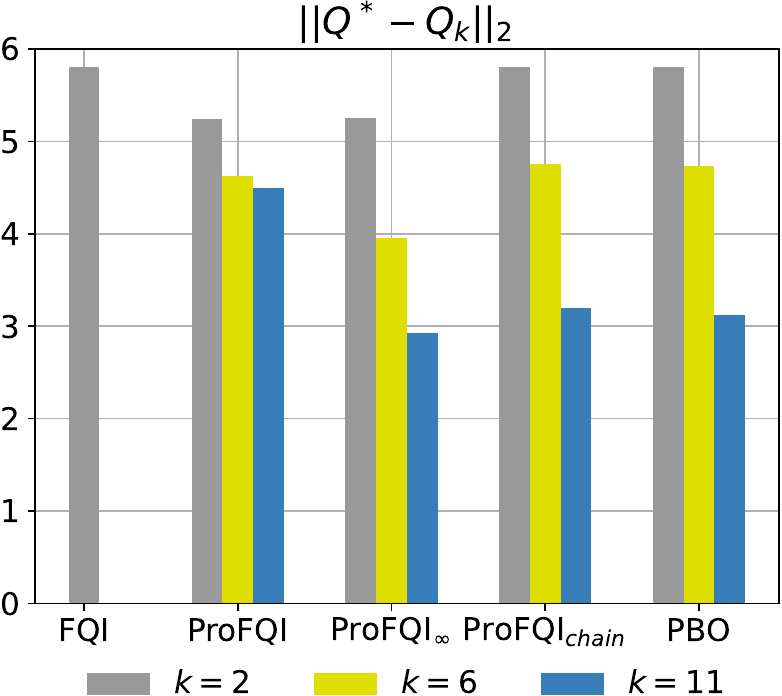}
        \caption{$K = 2$.}
        \end{center}
    \end{subfigure}
    \begin{subfigure}{0.33\textwidth}
        \begin{center}
        \includegraphics[scale=.35]{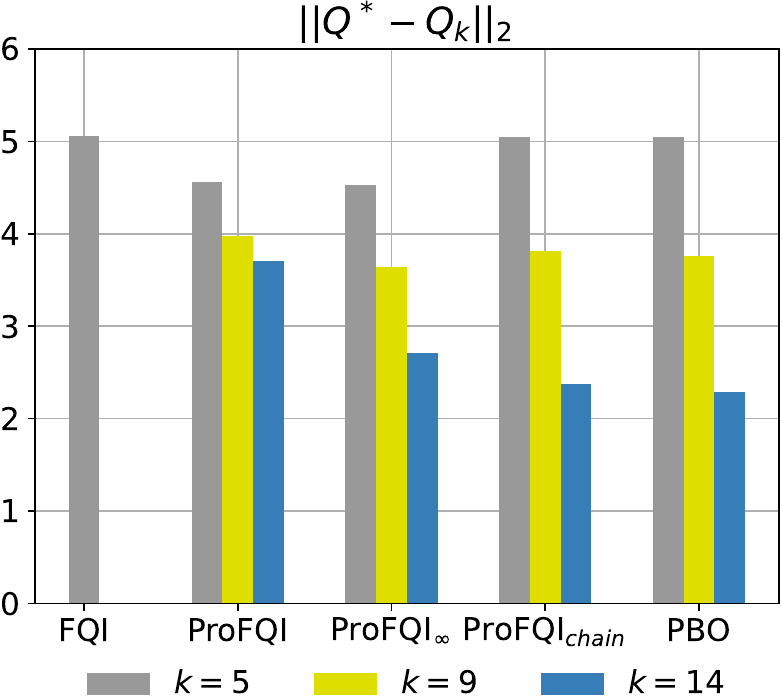}
        \caption{$K = 5$.}
        \end{center}
    \end{subfigure}
    \begin{subfigure}{0.33\textwidth}
        \begin{center}
        \includegraphics[scale=.35]{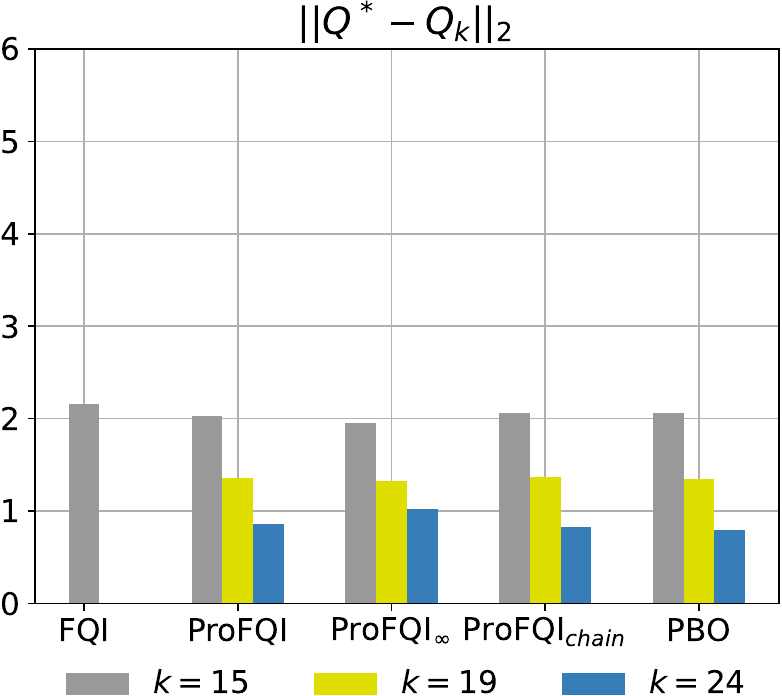}
        \caption{$K = 15$.}
        \end{center}
    \end{subfigure}
    \caption{$\ell_2$-norm of the difference between the optimal action-value function and the approximated action-value function on chain-walk. Here $K$ is the number of iterations included in the training loss (\ref{E:pbo_loss}) of PBO, while $k$ is the number of PBO applications after training. Results are averaged over $20$ seeds. Note that PBO enables using $k\geq K$ compared to FQI, which results in better convergence for increasing $k$ for each fixed $K \in \{2, 5, 15\}$.}\label{F:chain_walk_v}
\end{figure*}
\subsection{Finite Markov Decision Processes}\label{S:chain_walk}
\begin{figure}
    \centering
    \includegraphics[width=0.4\textwidth]{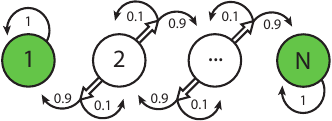}
    \caption{Test task: chain-walk from~\citet{munos2003error}. The reward is $0$ in all states except green states where it equals $1$.}
    \label{F:chain_walk}
\end{figure}
Let us consider finite state and action spaces of cardinality $N$ and $M$, respectively, and a tabular setting with $\Omega =\R^{N \cdot M}$. Given the use of tabular approximation, it is intuitive that each entry of the table can be modeled with a different single parameter, i.e., there is a bijection between $\mathcal{Q}_{\Omega}$ and $\Omega$, which allows us to write, for ease of notation, the parameters of the action-value function as $Q$ instead of $\vomega$.
\begin{proposition}
The PBO exists, and it is equal to the optimal Bellman operator
\begin{equation}
    \Lambda^*(Q) = R + \gamma P \max_{a\in\mathcal{A}}Q(\cdot,a).\label{E:PBO_finite}
\end{equation}
\end{proposition}
Note that PBO for finite MDPs is a $\gamma$-contraction mapping for the $L_\infty$-norm, like an optimal Bellman operator.
\begin{figure*}[]
    \centering
    \begin{subfigure}{0.33\textwidth}
        \begin{center}
        \includegraphics[scale=.35]{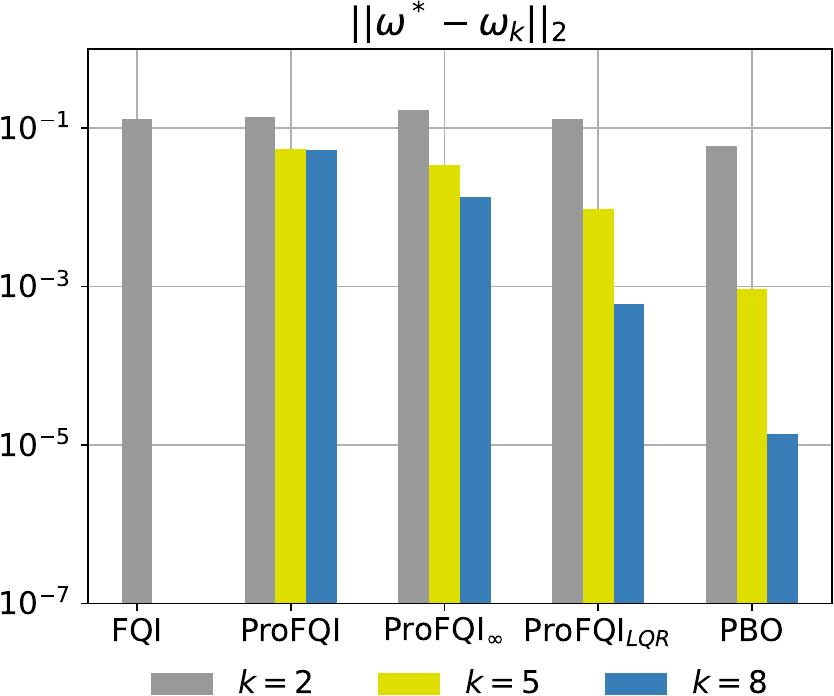}
        \subcaption{$K = 2$.}
        \label{F:lqr_v}
        \end{center}
    \end{subfigure}
    \begin{subfigure}{0.33\textwidth}
        \begin{center}
        \includegraphics[scale=.35]{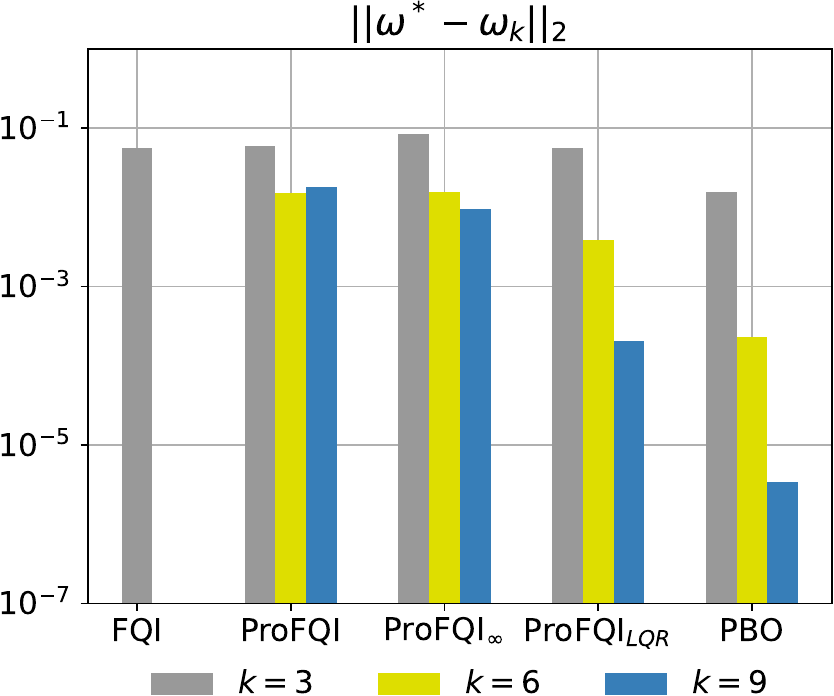}
        \subcaption{$K = 3$.}
        \label{F:lqr_n_b_analysis}
        \end{center}
    \end{subfigure}
    \begin{subfigure}{0.33\textwidth}
        \begin{center}
        \includegraphics[scale=.35]{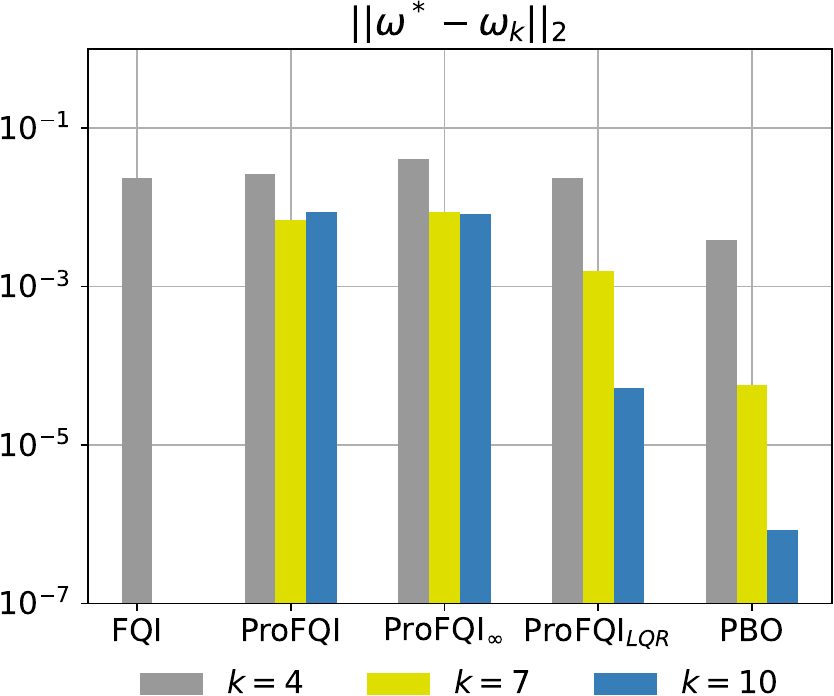}
        \caption{$K = 4$.}
        \end{center}
    \end{subfigure}
    \caption{$\ell_2$-norm of the difference between the optimal action-value function parameters and the estimated ones on LQR. Here $K$ is the number of iterations included in the training loss (\ref{E:pbo_loss}) of PBO, while $k$ is the number of PBO applications after training. Results are averaged over $20$ seeds. Similar to Figure~\ref{F:chain_walk_v}, PBO shows improved convergence for $k\geq K$ compared to FQI.}\label{F:lqr_w}
\end{figure*}
As an example of finite MDP, we consider the chain-walk environment in Figure~\ref{F:chain_walk}, with a chain of length $N=20$. We parameterize value functions as tables to leverage our theoretical results on finite MDPs. In Figure~\ref{F:chain_walk_v}, we show the $\ell_2$-norm of the difference between the optimal action-value function and the action-value function computed with FQI and ProFQI. We consider three different values of Bellman iterations, namely $K\in\lbrace 2, 5, 15 \rbrace$. For FQI, the $K$ iterations are the regular iterations where the empirical Bellman operator is applied to the current approximation of the action-value function. For ProFQI, $K$ is the number of iterations included in the PBO loss. This ensures a fair comparison, given that both methods have access to the same number of Bellman iterations. Once PBO is trained, we apply it for different numbers of iterations $k \geq K$, on given action-value function parameters. In Figure~\ref{F:chain_walk_v}, ProFQI uses a linear approximation of PBO trained with the loss~(\ref{E:pbo_loss}), ProFQI$_\infty$ uses a linear approximation of PBO trained with the loss~(\ref{E:pbo_loss_plus}), ProFQI$_\text{chain}$ uses the closed-form PBO in Equation~(\ref{E:PBO_finite}) considering $R$ and $P$ as unknown parameters to learn, and PBO indicates the use of the same closed-form solution assuming known $R$ and $P$. For the three variants of ProFQI, we observe that the approximation error decreases as the number of PBO iterations increases, evincing that PBO accurately emulates the true Bellman operator. In the case of $K=2$ and $K=5$, we see that ProFQI$_\infty$ and ProFQI$_\text{chain}$ obtain a better approximation of the action-value function compared to ProFQI, thanks to, respectively, the inclusion of fixed point in the loss and the use of the closed-form solution. Interestingly, in the case of $K=15$, ProFQI$_\infty$ obtains a slightly worse approximation than ProFQI. We explain this behavior with the fact that when the linear approximation is inadequate for modeling PBO, adding the fixed point in the loss could harm the estimate.
\subsection{Linear Quadratic Regulation}\label{S:lqr_experiment}
Now, we consider the continuous MDPs class of linear quadratic regulator~(LQR) with $\mathcal{S} = \mathcal{A} = \mathbb{R}$. The transition model $\mathcal{P}(s,a) = As + Ba$ is deterministic and the reward function $\mathcal{R}(s, a) = Qs^2 + 2 Ssa + Ra^2$ is quadratic, where $A$, $B$, $Q$, $S$ and $R$, are constants inherent to the MDP. We choose to parameterize the action-value functions with a $2$-dimensional parameter vector $\mathcal{Q}_{\Omega} = \{ (s, a) \mapsto G s^2 + 2 I sa + M a^2 | (G, I) \in \R^2 \}$ where $M$ is a chosen constant, for visualization purposes.
\begin{proposition}
PBO exists, and for any $\vomega \in \R^2$, its closed form is given by:\footnote{Under a mild assumption over the sample distribution, see in the Proofs section in the appendix.}
\begin{equation}
    \Lambda^* : \vomega=\begin{bmatrix}
        G \\
        I
    \end{bmatrix}
    \mapsto 
    \begin{bmatrix}
        Q + A^2 (G - \frac{I^2}{M}) \\
        S + AB (G - \frac{I^2}{M})
    \end{bmatrix}.\label{E:PBO_lqr}
\end{equation}
\end{proposition}
\begin{figure}
    \centering
    \includegraphics[width=\columnwidth]{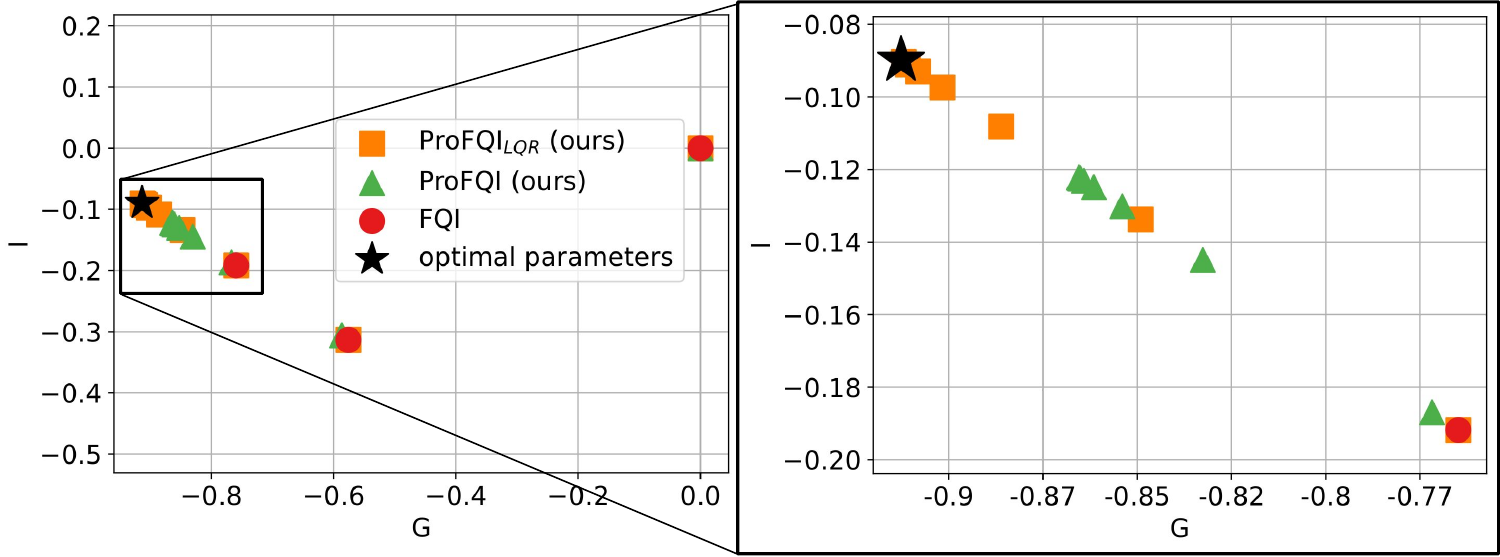}
    \caption{Behavior of FQI and ProFQI in the space of action-value function parameters $(G,I)$ for LQR. FQI is performed with $2$ iterations, and ProFQI uses a PBO trained with $K=2$, for the sake of fair comparison. Notably, ProFQI can leverage PBO by applying it for $8$ iterations, being able to get closer to the optimal parameters (black star) than FQI.}\label{F:lqr_2d_visualization}
\end{figure}
We leverage our theoretical analysis for LQR~\citep{bradtke1992reinforcement,pang2021robust} by parameterizing value functions accordingly, and we conduct a similar analysis to the one for chain-walk. This time, we evaluate the distance between the optimal action-value function parameters, which can be computed analytically, and the estimated ones. We use the closed-form solution obtained in Equation~\ref{E:PBO_lqr} assuming the parameters $(A,B,Q,S)$ known (PBO), and unknown (ProFQI$_\text{LQR}$). Figure~\ref{F:lqr_w} confirms the pattern observed on the chain-walk. ProFQI and ProFQI$_\infty$ obtain a better approximation than FQI, which is reasonably worse than ProFQI$_\text{LQR}$ and PBO. We also observe that ProFQI$_\text{LQR}$ obtains a significantly better approximation than the other variants for a large number of iterations (blue bars), confirming the advantages of exploiting the closed-form solution of PBO for finite MDPs. We additionally show the sequence of parameters corresponding to the iterations of FQI, ProFQI$_\text{LQR}$, and ProFQI, in Figure~\ref{F:lqr_2d_visualization}. The training is done with $K = 2$. The sequence starts from the chosen initial parameters $(G,I)=(0,0)$ for all algorithms and proceeds towards the optimal parameters, which are computed analytically. Both ProFQI and ProFQI$_\text{LQR}$ apply the PBO learned after the $K=2$ iterations, for $8$ iterations; thus, the sequence of parameters for both algorithms is composed of $8$ points each, while FQI has $2$. It is clear that the parameters found by FQI are considerably further to the target parameters than the ones found by both ProFQI and ProFQI$_\text{LQR}$. In particular, the latter gets the closest to the target, in line with the results in Figure~\ref{F:lqr_w}, again evincing the accuracy of the learned PBO and the benefit of performing multiple applications of it, as expected from Theorem~\ref{Th:error_propagation}.

\begin{figure}
    \centering
    \begin{subfigure}{.49\columnwidth}
        \begin{center}
            \includegraphics[width=\columnwidth]{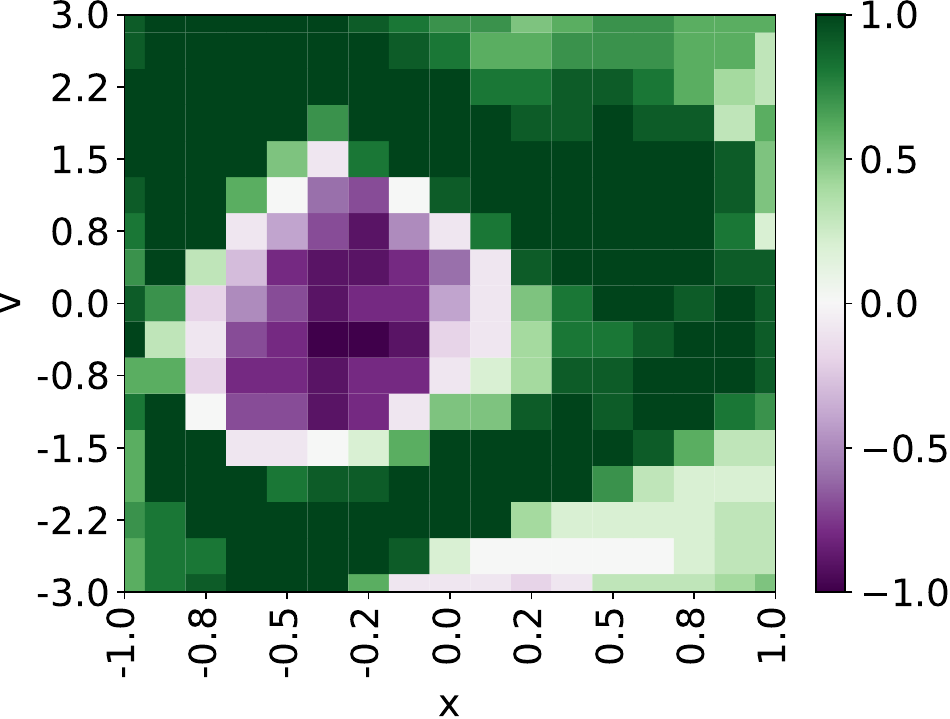}
            \caption{ProFQI (ours).}
            \label{F:car_on_hill_policy_pbo}
        \end{center}
    \end{subfigure}
    \hspace{-.35cm}
    \begin{subfigure}{.49\columnwidth}
        \begin{center}
            \includegraphics[width=.85\columnwidth]{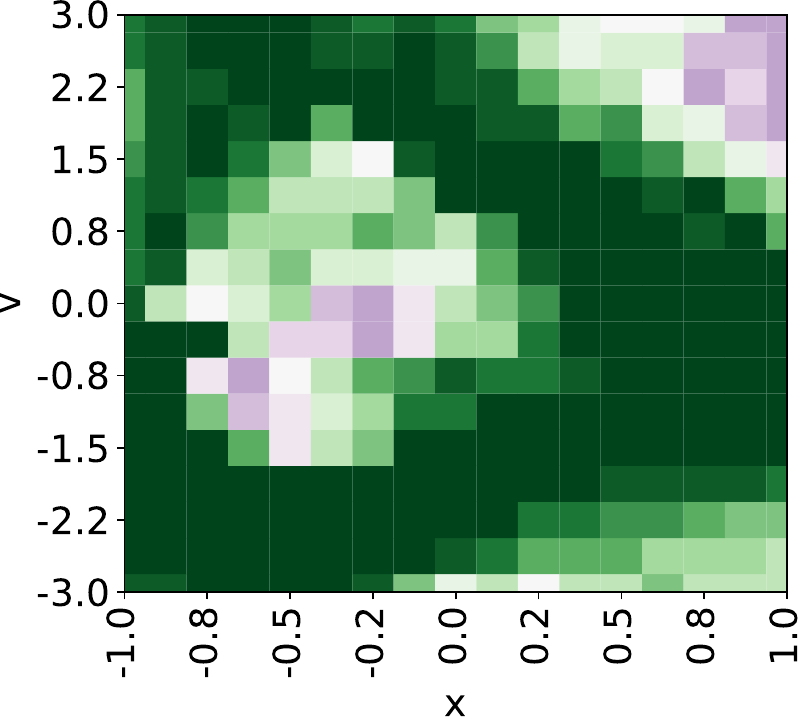}
            \caption{FQI.}
            \label{F:car_on_hill_policy_fqi}
        \end{center}
    \end{subfigure}
    \caption{Policies on car-on-hill using $K=9$ and $9$ application of PBO, averaged over $20$ seeds. The purple and green colors refer to the two discrete actions (move left or right). Our ProFQI approximates the optimal policy more closely.}
    \label{F:car_on_hill_policies}
\end{figure}

\begin{figure*}[t]
    \centering
    \begin{subfigure}{0.33\textwidth}
        \begin{center}
        \includegraphics[scale=.35]{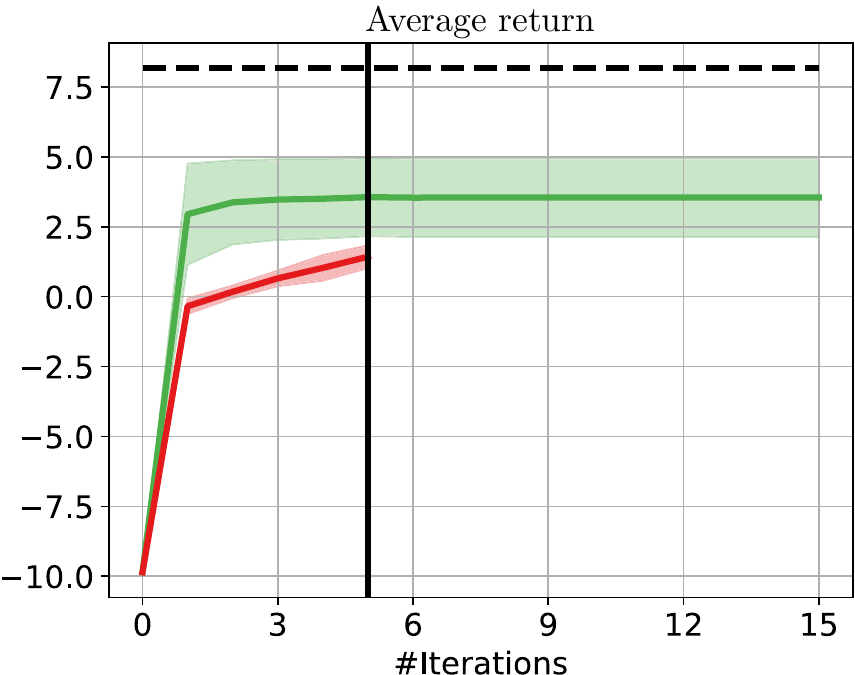}
        \subcaption{$K = 5$.}
        \end{center}
    \end{subfigure}
    \begin{subfigure}{0.33\textwidth}
        \begin{center}
        \includegraphics[scale=.35]{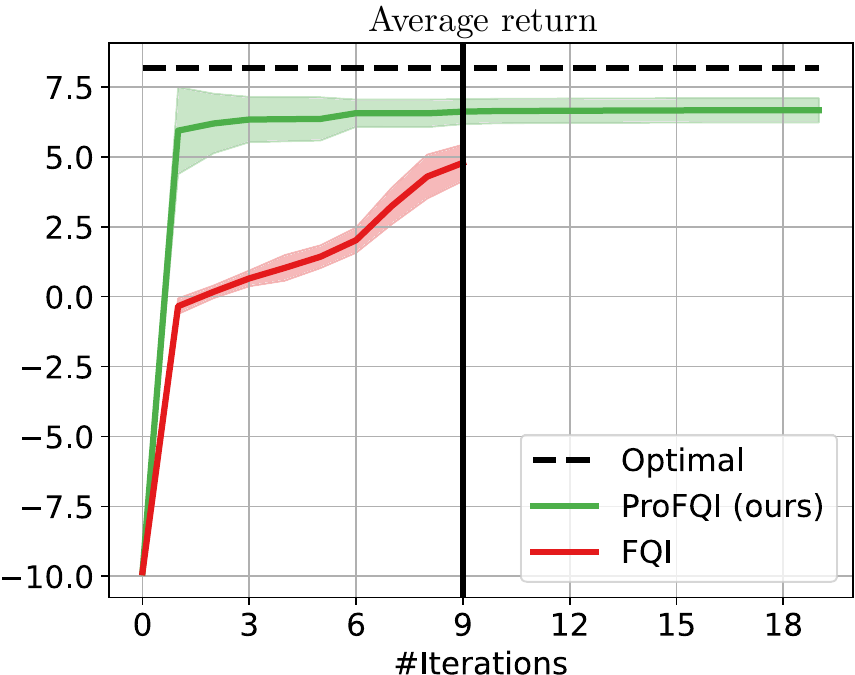}
        \subcaption{$K = 9$.}
        \end{center}
    \end{subfigure}
    \begin{subfigure}{0.33\textwidth}
        \begin{center}
        \includegraphics[scale=.35]{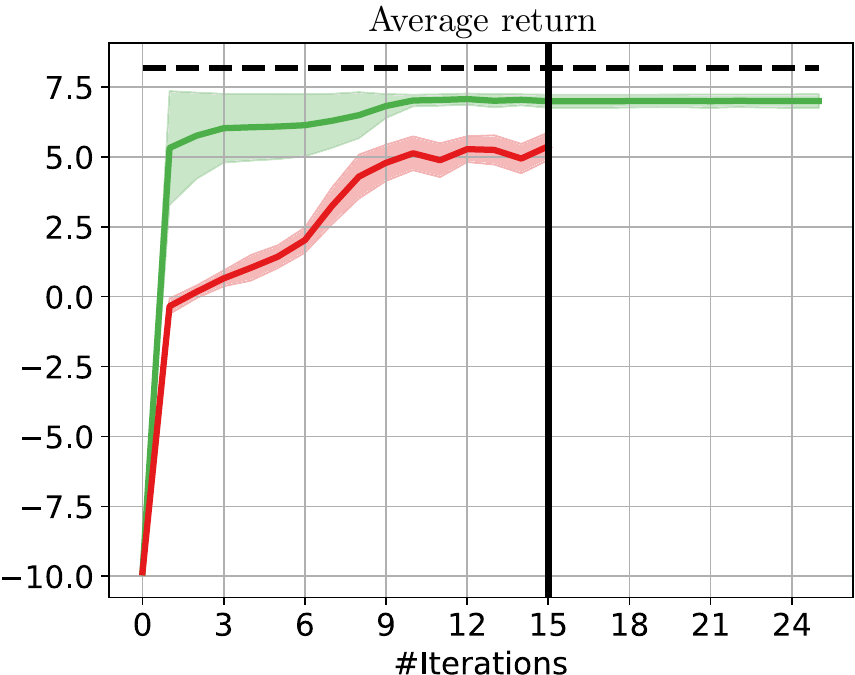}
        \caption{$K = 15$.}
        \end{center}
    \end{subfigure}
    \caption{Discounted cumulative reward on car-on-hill. The black vertical line indicates the number of iterations considered for FQI and the number of iterations $K$ considered to train PBO. Results are averaged over $20$ seeds with $95\%$ confidence intervals. Note that ProFQI achieves higher performance than FQI and remains stable after $K$ iterations.}\label{F:car_on_hill_J}
\end{figure*}
\begin{figure}
    \centering
    \begin{subfigure}{0.45\columnwidth}
        \begin{center}
        \includegraphics[width=\columnwidth]{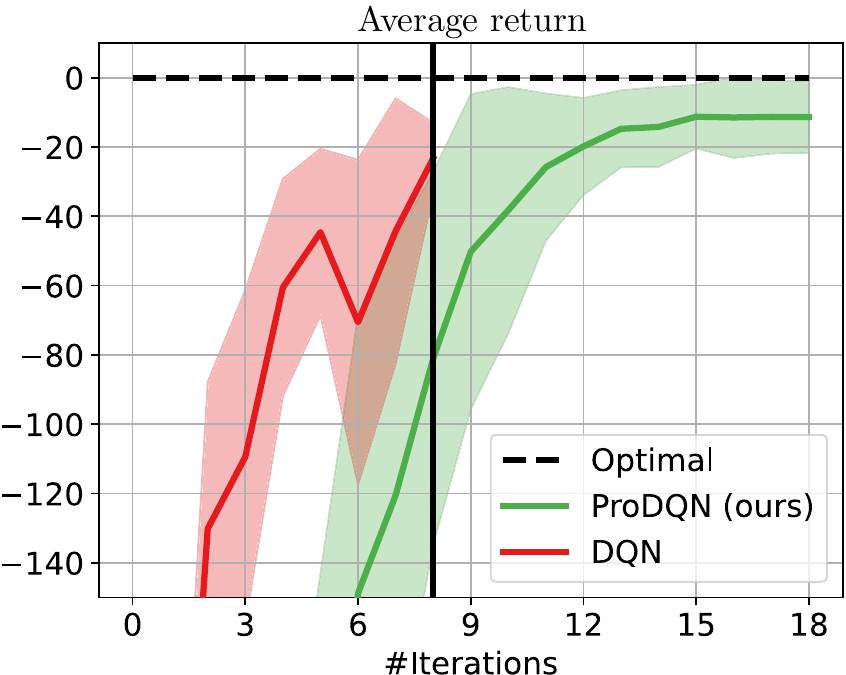}
        \subcaption{Bicycle balancing.}\label{F:bicycle_online_J}
        \end{center}
    \end{subfigure}
    \hspace{.05cm}
    \begin{subfigure}{0.45\columnwidth}
        \begin{center}
        \includegraphics[width=\columnwidth]{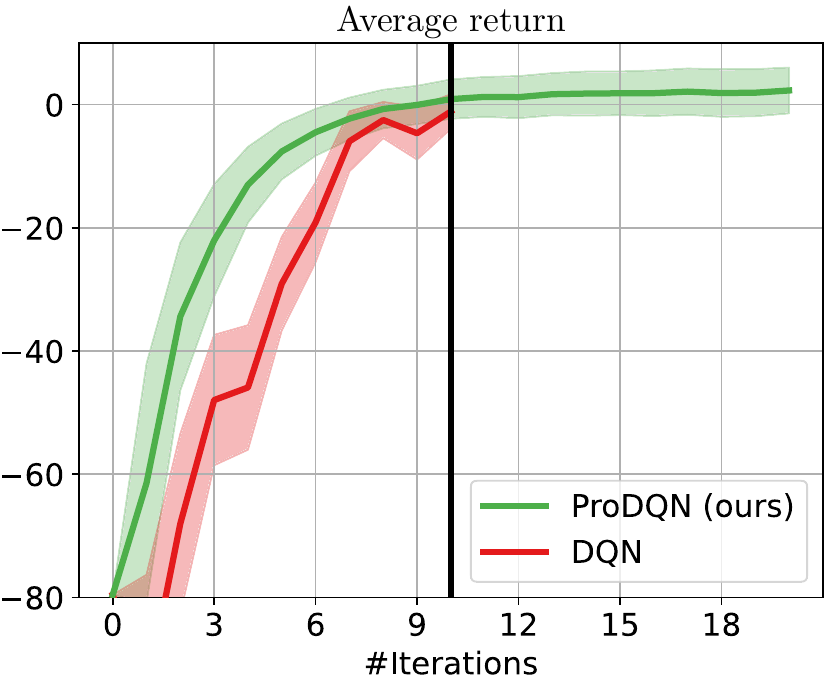}
        \subcaption{Lunar lander.}\label{F:lunar_lander_J}
        \end{center}
    \end{subfigure}
    \caption{Discounted cumulative reward on bicycle and lunar lander. The black vertical line indicates the number of target network updates performed by DQN and the number of iterations $K$ considered to train PBO. Results are averaged over $40$ seeds with $95\%$ confidence intervals.}\label{F:pro_dqn_J}
\end{figure}

\section{Experiments}\label{S:experiments}
We consider both ProFQI and ProDQN, comparing their performance with their regular counterparts\footnote{The code is available at \url{https://github.com/theovincent/PBO}}. To handle the increased complexity of the input space of the considered problems, we leverage neural network regression to model our PBO. We consider an offline setting, where we use ProFQI on car-on-hill~\citep{ernst2005fqi}, and an online setting, where we use ProDQN on bicycle balancing~\citep{randlov1998learning}, and lunar lander~\cite{brockman2016openai}. We want to answer the following research question: \emph{does PBO enable moving toward the fixed point more effectively than the empirical Bellman operator?}

\subsection{Projected Fitted $Q$-Iteration}
We initially evaluate ProFQI on the \textit{car-on-hill} problem. As done in~\citet{ernst2005fqi}, we measure performance by generating roll-outs starting from different states on a grid of size $17 \times 17$, and accounting for the fact that the dataset $\mathcal{D}$ does not contain every starting state of the grid, by weighting the obtained performance from each starting state by the number of times it occurs in the dataset (see appendix). First, the benefit of PBO is observable in the quality of the policy computed by FQI and ProFQI (Figure~\ref{F:car_on_hill_policies}). After only $9$ training iterations, ProFQI obtains a policy that is very close to the optimal one of car-on-hill (see the well-known shape of the optimal car-on-hill policy in Figure~\ref{F:car_on_hill_policy_pbo}~\cite{ernst2005fqi}), while FQI is significantly more inaccurate. The consequences of this are reflected in the performance shown in Figure~\ref{F:car_on_hill_J}, that are obtained with FQI and ProFQI for three different values of Bellman iterations $K$ (black vertical line). Again, for FQI, $K$ is the number of regular iterations consisting of an application of the empirical Bellman operator and the projection step; for ProFQI, $K$ is the number of applications of PBO that are used in the training loss (Equation~\ref{E:pbo_loss}). The iterations on the $x$-axis in Figure~\ref{F:car_on_hill_J} are the regular iterations for FQI and the applications of the trained PBO for ProFQI. Thus, the iterations on the right side of the line can only be reached with ProFQI. The purpose of this analysis is to evaluate the different quality of trajectories toward the fixed point obtained by the empirical Bellman operator in FQI, and the trained PBO in ProFQI, for the same amount of transition samples. We observe that ProFQI obtains better performance than FQI consistently. This evinces that PBO learns an accurate estimate of the true Bellman operator, which allows obtaining a satisfactory action-value function faster than the standard empirical Bellman operator used by FQI.
It is interesting to note that the performance of ProFQI remains stable for subsequent applications of PBO after the number of iterations used for training.

\subsection{Projected Deep $Q$-Network}
We also evaluate PBO in an online setting, using our ProDQN algorithm and comparing against DQN~\cite{mnih2015human}. We consider a \textit{bicycle balancing}~\cite{randlov1998learning} problem and the \textit{lunar lander} environment~\cite{brockman2016openai}. We set the number of Bellman iterations $K=8$ for bicycle balancing and $K=10$ for the lunar lander. Similar to the offline setting, $K$ is the number of updates of the target network for DQN, and the number of applications of PBO in the training loss (Equation~\ref{E:pbo_loss}) for ProDQN. Due to the need to collect new samples while learning a policy, training PBO in an online setting needs a slightly different treatment than the offline case. Recalling Algorithm~\ref{A:algorithm}, we point out that new samples are collected after each gradient descent step, by using the action-value function obtained after $K$ applications of the current PBO. We find this choice to work well in practice; however, we can envision multiple possibilities for effective exploration strategies based on PBO, that we postpone to future works.

Figure~\ref{F:pro_dqn_J} shows the discounted cumulative reward collected by DQN and ProDQN on both bicycle balancing and lunar lander, for different numbers of iterations. For DQN, each iteration corresponds to an update of the target network, while for ProDQN it indicates an application of the trained PBO. Figure~\ref{F:bicycle_online_J} shows that contrary to car-on-hill, on the bicycle balancing task ProDQN improves slower than DQN, but it keeps increasing performance after $K$ iterations thanks to PBO. This behavior illustrates that PBO can be used after the training to move the parameters in a favorable direction (see Figure \ref{F:pbo} (blue trajectory)). Similarly, Figure~\ref{F:lunar_lander_J} shows that, on lunar lander, ProDQN keeps increasing ever so slightly, but it is also stronger than DQN overall.

\section{Discussion and Conclusion}\label{S:conclusion}
We introduced the novel idea of an operator that directly maps parameters of action-value functions to others, as opposed to the regular Bellman operator that requires costly projections steps onto the space of action-value functions. This operator called the projected Bellman operator can generate a sequence of parameters that can progressively approach the ones of the optimal action-value function. We formulated an optimization problem and an algorithmic implementation to learn PBO in offline and online RL. One limitation of our method in its current form is that, given that the size of input and output spaces of PBO depends on the number of parameters of the action-value function, it is challenging to scale to problems that learn action-value functions with deep neural networks with millions of parameters~\cite{mnih2015human}. Nevertheless, recent advances in deep learning methods for learning operators~\cite{kovachki2021neural,kovachki2023neural} can provide a promising future direction.

\section{Acknowledgments}
This work was funded by the German Federal Ministry of Education and Research (BMBF) (Project: 01IS22078). This work was also funded by Hessian.ai through the project ’The Third Wave of Artificial Intelligence – 3AI’ by the Ministry for Science and Arts of the state of Hessen and by the grant “Einrichtung eines Labors des Deutschen Forschungszentrum für Künstliche Intelligenz (DFKI) an der Technischen Universität Darmstadt”.
This paper was also supported by FAIR (Future Artificial Intelligence Research) project, funded by the NextGenerationEU program within the PNRR-PE-AI scheme (M4C2, Investment 1.3, Line on Artificial Intelligence).
\bibliography{aaai24}

\newpage
\appendix
\onecolumn

\section{Projected Bellman Operator in Low-Rank Markov Decision Processes}\label{A:low_rank_mdps}
Low-rank MDPs is a class of problems with two feature maps $\vsigma : \mathcal{S} \times \mathcal{A} \to \mathbb{R}^d$ and $\vmu : \mathcal{S} \to \mathbb{R}^d$, such that $\mathcal{P}(s'| s, a) = \langle \vsigma(s, a) , \vmu(s') \rangle$ and $\mathcal{R}(s, a) = \langle \vsigma(s, a) , \vtheta \rangle$, for all $(s, a, s') \in \mathcal{S} \times \mathcal{A} \times \mathcal{S}$ and for $\vtheta \in \mathbb{R}^d$~\citep{agarwal2020flambe,sekhari2021agnostic}. We assume, without loss of generality, that for all $(s, a) \in \mathcal{S} \times \mathcal{A}, \lVert \vsigma(s, a) \rVert_1 \leq 1$ and $\max\lbrace\lVert\vmu(s) \rVert_1, \lVert \vtheta \rVert_1\rbrace \leq \sqrt{d}$. The space of action-value functions is set as the space of linear functions in the parameters, i.e., $\mathcal{Q}_{\Omega} = \lbrace \langle \vsigma(\cdot, \cdot) , \vomega \rangle \vert \vomega \in \mathbb{R}^d \rbrace$.
\begin{proposition}
In the case of continuous state and action spaces and for $\vomega \in \R^d$, the PBO is
\begin{equation}
    \Lambda^*(\vomega) = \vtheta + \gamma \int_{\mathcal{S}} \max_{a' \in \mathcal{A}} \langle \vsigma(s', a') | \vomega \rangle  \vmu(s') \mathrm{d}s'.
\end{equation}
\end{proposition}
The closed form is again a $\gamma$-contraction mapping assuming that the MDP has a latent variable representation, that the state and action spaces are compact and that the feature map $\vsigma$ is continuous.

\section{Proofs}\label{A:proofs}
\subsection{Closed-Form of PBO for a Finite MDP}
\begin{proof}
We compute the optimal Bellman iteration over a table $Q \in \R^{N \cdot M}$
\begin{align}
    \Gamma^*Q(s, a)
    &= r(s, a) + \gamma \E_{s' \sim p(\cdot | s, a)} \left[ \max_{a' \in \mathcal{A}} [Q(s', a')] \right] \nonumber \\
    &= r(s, a) + \gamma \sum_{s'} p(s'| s, a) \left[ \max_{a' \in \mathcal{A}} [Q(s', a')] \right] \nonumber \\
    &= \left(R + \gamma P \max_{a' \in \mathcal{A}} Q(\cdot, a') \right)(s, a)
\end{align}
where $P \in \R^{N \cdot M \times N}$ is the transition probability matrix of the environment. From these equations, the operator $Q \mapsto R + \gamma P \max_{a' \in \mathcal{A}} Q(\cdot, a')$, evaluated on the objective function from the definition of PBO in Equation \ref{E:opt_pbo}, yields zero error. This means that we have found the PBO in closed form.
\end{proof}

\subsection{Closed-Form of PBO for LQR MDP}
\begin{proof}
We assume that the distribution over the samples is a discrete uniform distribution over $\mathcal{S} \times \mathcal{A}$ centered on zero in both directions. With this assumption, the optimization problem in Equation \ref{E:opt_pbo} is equivalent to:
\begin{equation*}
    \argmin_{\Lambda : \Omega \to \Omega} \mathbb{E}_{\vomega \sim \nu} \left[ \sum_{(s, a) \in \mathcal{\bar{S}} \times \mathcal{\bar{A}}} \left(\Gamma^*Q_{\vomega}(s,a) - Q_{\Lambda(\vomega)}(s,a)\right)^2 \right]
\end{equation*}
where $\mathcal{\bar{S}} \times \mathcal{\bar{A}}$ is the set of all possible state-action pairs that can be drawn by the distribution of samples. \\

Let $\Lambda$ be an operator on $\Omega = \R^2$, $\vomega = (G, I) \in \Omega$ and $(s, a) \in \mathcal{\bar{S}} \times \mathcal{\bar{A}}$. We note the first and second components of $\Lambda(
\omega)$, $\Lambda_G(\omega)$ and $\Lambda_I(\omega)$, we have:
\begin{equation*}
    \Gamma^*Q_{\vomega}(s, a) - Q_{\Lambda(\vomega)}(s, a)
    = \begin{bmatrix} - s^2 & -2 sa \end{bmatrix} \begin{bmatrix}
        \Lambda_G(\omega) \\
        \Lambda_I(\omega)
    \end{bmatrix} + \Gamma^*Q_{\vomega}(s, a) - M a^2.
\end{equation*}
We note $Z(s, a) = \begin{bmatrix} - s^2 & -2 sa \end{bmatrix}$, $X = 
\begin{bmatrix} \Lambda_G(\omega) & \Lambda_I(\omega) \end{bmatrix}^T$ and $b(s, a) = \Gamma^*Q_{\vomega}(s, a) - M a^2$. By summing over the samples we get:
\begin{equation*}
    \sum_{(s, a) \in \mathcal{\bar{S}} \times \mathcal{\bar{A}}} \left(\Gamma^*Q_{\vomega}(s,a) - Q_{\Lambda(\vomega)}(s,a)\right)^2 = \sum_{(s, a) \in \mathcal{\bar{S}} \times \mathcal{\bar{A}}} \left( Z(s, a) X + b(s, a) \right)^2 = || Z X + b ||^2_2
\end{equation*}
where $Z = \begin{bmatrix} \hdots \\ Z(s, a) \\ \hdots \end{bmatrix}$ and $b = \begin{bmatrix} \hdots & b(s, a) & \hdots \end{bmatrix}^T$. \\
Minimizing $|| Z X + b ||^2_2$ over $X \in \Omega$ will bring us to the minimizer of the optimization problem for any parameter distribution $\nu$. To minimize this quantity, we need to investigate the matrix 
\begin{equation*}
    Z^T Z =
    \begin{bmatrix}
        \sum_{s} s^4 & 2 \sum_{s, a} s^3 a \\
        2 \sum_{s} s^3 a & 2 \sum_{s, a} s^2 a^2
    \end{bmatrix} =
    \begin{bmatrix}
        \sum_{s} s^4 & 0 \\
        0 & \sum_{s, a} s^2 a^2
    \end{bmatrix}.
\end{equation*}
The last equality comes from the fact that the sample distribution is symmetric along $\mathcal{\bar{S}}$ and $\mathcal{\bar{A}}$. $Z^T Z$ is positive definite so $\min_{X \in \Omega} || Z X + b ||^2_2$ has a unique minimizer: $\Lambda^*(\omega) = -(Z^T Z)^{-1} Z^T b$. \\

Let us now rewrite $b$. The optimal Bellman iteration over $Q_{\vomega}$ is $\Gamma^*Q_{\vomega}(s, a) = r(s, a) + \max_{a'} Q_{\vomega}(s', a')$. $M$ is chosen negative so that the function $a' \mapsto Q_{\vomega}(s', a') = G \cdot s'^2 + 2 I \cdot s' a' + M \cdot a'^2$ has a unique maximizer of equation $\nicefrac{-I}{M} \cdot s'$. This makes
\begin{equation*}
    \max_{a'} Q_{\vomega}(s', a') = Q_{\vomega}(s', -\frac{I}{M} \cdot s')
    = G \cdot s'^2 - 2 \frac{I^2}{M} \cdot s'^2 + \frac{I^2}{M} \cdot s'^2
    = (G - \frac{I^2}{M}) \cdot s'^2.
\end{equation*}
By inserting it into the Bellman equation, we get
\begin{align*}
    b(s, a) 
    &= \Gamma^*Q_{\vomega}(s, a) - M a^2 \\
    &= r(s, a) + (G - \frac{I^2}{M}) \cdot s'^2 - M a^2 \\
    &= Q \cdot s^2 + 2 S \cdot sa + R \cdot a^2 + (G - \frac{I^2}{M}) \cdot s'^2 - M a^2 \\
    &= \left(Q + A^2 (G - \frac{I^2}{M}) \right) \cdot s^2 + 2 \left( S + AB (G - \frac{I^2}{M}) \right) \cdot sa + \left(R + B^2 (G - \frac{I^2}{M}) - M \right) \cdot a^2 \\
    &= \begin{bmatrix}
        s^2 & 2 sa & a^2
    \end{bmatrix} \cdot
    \begin{bmatrix}
        Q + A^2 (G - \frac{I^2}{M}) - G \\
        S + AB (G - \frac{I^2}{M}) - I \\
        R + B^2 (G - \frac{I^2}{M}) - M
    \end{bmatrix}.
\end{align*}
This means that 
\begin{equation*}
    \Lambda^*(\omega) = -(Z^T Z)^{-1} Z^T J     
    \begin{bmatrix}
        Q + A^2 (G - \frac{I^2}{M}) \\
        S + AB (G - \frac{I^2}{M})\\
        R + B^2 (G - \frac{I^2}{M}) - M
    \end{bmatrix}
\end{equation*}
where $J = \begin{bmatrix}
    & \hdots & \\
    s^2 & 2 sa & a^2 \\
    & \hdots &
\end{bmatrix}$. \\
From the fact that $Z^T J = \begin{bmatrix}
    -\sum_{s} s^4 & 0 & 0\\
    0 & \sum_{s, a} s^2 a^2 & 0
\end{bmatrix}$, we have $-(Z^T Z)^{-1} Z^T J = \begin{bmatrix}
    1 & 0 & 0\\
    0 & 1 & 0
\end{bmatrix}$, thus $\Lambda^*(\omega) = \begin{bmatrix}
        Q + A^2 (G - \frac{I^2}{M}) \\
        S + AB (G - \frac{I^2}{M})
    \end{bmatrix}$.

\end{proof}

\paragraph{Remarks} With the assumption on the distribution of samples, the PBO can also be understood in a geometrical way. It projects the parameters along a line of direction vector $\begin{bmatrix} A^2 & AB \end{bmatrix}^T$ with an offset $\begin{bmatrix} Q & S \end{bmatrix}^T$. The iterations correspond to a non-linear transformation of the coefficient ($G - \nicefrac{I^2}{M}$) in front of the direction vector. This also means that the fixed point, i.e. the optimal parameters are also on this line.

\subsection{Closed-Form of PBO for a Low-Rank MDP}
\begin{proof}
The proof considers continuous unbounded state-action spaces. For $Q_{\vomega} \in \mathcal{Q}_{\Omega}$, $\vomega$ the vector representing $Q_{\vomega}$, $\max_{a \in \sA} Q_{\vomega}(s, a)$ is well defined (here $\max$ might not be attained, it should be interpreted as a supremum). We have $|Q_{\vomega}(s, a)| = | \langle \vsigma(s, a) | \vomega \rangle | \leq || \vsigma(s, a) || \cdot ||\vomega|| \leq ||\vomega||$, thus $\max_{a \in \sA} Q_{\vomega}(s, a) < \infty$. Then, we write the optimal Bellman iteration on the function $Q_{\vomega}$ as
\begin{align*}
    \Gamma^* &Q_{\vomega}(s, a) 
    = r(s, a) + \gamma \E_{s' \sim p(\cdot | s, a)} \left[ \max_{a' \in \sA} Q_{\vomega}(s', a') \right] \\
    &= r(s, a) + \gamma \int_{s'} \max_{a'} \langle \vsigma(s', a') | \vomega \rangle  p(s'| s, a) \mathrm{d}s' \\
    &= \langle \vsigma(s, a) | \vtheta \rangle + \gamma \int_{s'} \max_{a'} \langle \vsigma(s', a') | \vomega \rangle  \langle \vsigma(s, a) | \vmu(s') \rangle \mathrm{d}s' \text{ from the definition of the transition} \\ &\text{probabilities.}\\
    &= \langle \vsigma(s, a) | \vtheta + \gamma \int_{s'} \max_{a'} \langle \vsigma(s', a') | \vomega \rangle  \vmu(s') \mathrm{d}s' \rangle \text{ since the scalar product is linear in its second variable.}\\
\end{align*}
This derivation shows that the operator $\vomega \mapsto \vtheta + \gamma \int_{s'} \max_{a'} \langle \vsigma(s', a') | \vomega \rangle  \vmu(s') \mathrm{d}s'$ minimizes the objective function presented in the definition of PBO in Equation \ref{E:opt_pbo} since its yields zero error. This operator is the PBO for a low-rank MDP.
\end{proof}

\subsection{PBO Is a $\gamma$-Contraction Mapping for a Low-Rank MDP}
\begin{proof}
We now assume that the MDP has a latent variable representation (i.e. $\forall i \in \{1, \hdots, d\}, \vsigma_i(\cdot, \cdot)$ and $\vmu_i(\cdot)$ are probability distribution), that the state and action spaces are compact and that the feature map $\sigma$ is continuous. This proof was inspired by the proof of Proposition 6.2.4 in \citet{puterman2014markov}. Considering $\vomega, \vomega' \in \R^d$, we have
\begin{align*}
    || \Lambda^*&(\vomega) - \Lambda^*(\vomega') ||_{\infty} 
    = \gamma || \int_{s'} \left( \max_{a'} \langle \vsigma(s', a') | \vomega \rangle - \max_{a'} \langle \vsigma(s', a') | \vomega' \rangle \right)  \vmu(s') \mathrm{d}s' ||_{\infty} \\
    &= \gamma \max_{i \in \{1, \hdots, d \}} \left| \int_{s'} \left( \max_{a'} \langle \vsigma(s', a') | \vomega \rangle - \max_{a'} \langle \vsigma(s', a') | \vomega' \rangle \right) \vmu_i(s') \mathrm{d}s' \right| \\
    &\leq \gamma \max_{i \in \{1, \hdots, d \}} \int_{s'} \left| \max_{a'} \langle \vsigma(s', a') | \vomega \rangle - \max_{a'} \langle \vsigma(s', a') | \vomega' \rangle \right| \vmu_i(s') \mathrm{d}s' \text{ since each element of } \vmu(\cdot) \text{ is positive.}\\
    &\leq \gamma \max_{i \in \{1, \hdots, d \}} \max_{s'} \left| \max_{a'} \langle \vsigma(s', a') | \vomega \rangle - \max_{a'} \langle \vsigma(s', a') | \vomega' \rangle \right| \text{ since each } \vmu_i(\cdot) \text{ is a probability distribution.}\\
    &\text{the maximum is reached because the state space is compact and the function } s' \mapsto \max_{a'} \langle \vsigma(s', a') | \vomega' \rangle \text{ is continuous.}\\
    &\leq \gamma \max_{s'} \left| \max_{a'} \langle \vsigma(s', a') | \vomega \rangle - \max_{a'} \langle \vsigma(s', a') | \vomega' \rangle \right|\\
    &\leq \gamma || \vomega - \vomega' ||_{\infty}
\end{align*}
The last inequality comes from the following. Since the action space is compact and the feature function $\vsigma$ is continuous, there exists an action $a_{s', \vomega}$ such that $a_{s', \vomega} = \argmax_{a'} \langle \vsigma(s', a') | \vomega \rangle$ from the Weierstrass extreme value theorem. Similarly, we note $a_{s', \vomega'} = \argmax_{a'} \langle \vsigma(s', a') | \vomega' \rangle$. Then, we have
\begin{align*}
    \max_{a'} \langle &\vsigma(s', a') | \vomega \rangle - \max_{a'} \langle \vsigma(s', a') | \vomega' \rangle\\
    &\leq \langle \vsigma(s', a_{s', \vomega}) | \vomega \rangle - \langle \vsigma(s', a_{s', \vomega}) | \vomega' \rangle\\
    &= \langle \vsigma(s', a_{s', \vomega}) | \vomega - \vomega' \rangle\\
    &\leq || \vomega - \vomega' ||_{\infty} \text{ since } || \vsigma(s', a_{s', \vomega}) ||_1 \leq 1.\\
\end{align*}
By symmetry, we have $\left| \max_{a'} \langle \vsigma(s', a') | \vomega \rangle - \max_{a'} \langle \vsigma(s', a') | \vomega' \rangle \right| \leq || \vomega - \vomega' ||_{\infty}$, hence the result.
\end{proof}

\section{Details of the Empirical Analysis}\label{A:experiments}
We provide details of the experimental setting. Table \ref{T:parameters_offline} and \ref{T:parameters_online} summarize the values of all parameters appearing in the experiments. FQI, DQN, ProFQI, and ProDQN use Adam optimizer~\citep{kingma2015adam} with a linear annealing learning rate. For FQI, the optimizer is reset at each iteration. The set of parameters $\mathcal{W}$ is generated by sampling from a truncated normal distribution. When we use action-value functions with a neural network that has more than one hidden layer, the last one is initialized with zeros. This way, the output of a neural network parameterized by any element of $\mathcal{W}$ is zero which makes the reward easier to learn. Among the parameters in $\mathcal{W}$, one is chosen to be the initial parameters used by FQI or DQN. For all the methods, the metrics have been constructed starting from the initial parameters, this is why, all the plots showing the performances share the same $J_0$. As Table \ref{T:parameters_offline} and \ref{T:parameters_online} show, the initial parameters of PBOs are always taken small enough so that applying PBO multiple times does not lead to diverging outputs.

\subsection{Offline Experiments}

\begin{table}
\centering
\caption{Summary of all parameters used in the offline experiments.}\label{T:parameters_offline}
\begin{tabular}{|| c | c || c | c | c | c ||} 
    \hline
    \multicolumn{2}{|| c ||}{} & Chain-walk & LQR & Car-on-hill & Bicycle \\ 
    \hline \hline
    \multicolumn{2}{|| c ||}{horizon} & $+\infty$ & $+\infty$ & $100$ & $50.000$ \\
    \hline
    \multicolumn{2}{|| c ||}{$\gamma$} & $0.9$ & $1$ & $0.95$ & $0.99$ \\
    \hline
    \multicolumn{2}{|| c ||}{\#$\mathcal{D}$} & $400$ & $121$ & $5.500$ & $70.000$ \\
    \hline
    \multicolumn{2}{|| c ||}{batch size on $\mathcal{D}$} & $20$ & $121$ & $500$ & $1.000$ \\
    \hline \hline
    \multirow{4}{*}{FQI} 
    & \#fitting steps & $400$ & $800$ & $1.200$ & $1.200$ \\
    \cline{2-6}
    & \#patience steps & $100$ & $100$ & $30$ & $7$ \\
    \cline{2-6}
    & starting learning rate & $10^{-2}$ & $10^{-2}$ & $10^{-3}$ & $5 \times 10^{-3}$ \\
    \cline{2-6}
    & ending learning rate & $10^{-5}$ & $10^{-5}$ & $5 \times 10^{-7}$ & $10^{-4}$ \\
    \hline  \hline
    \multirow{8}{*}{ProFQI} 
    & \#$\mathcal{W}$ & $100$ & $5$ & $30$ & $50$ \\
    \cline{2-6}
    & batch size on $\mathcal{W}$ & $100$ & $5$ & $30$ & $25$ \\
    \cline{2-6}
    & \#epochs & $1.000$ & $1.000$ & $1.000$ & $500$ \\
    \cline{2-6}
    & \#training steps & $5$ & $4$ & $10$ & $20$ \\
    \cline{2-6}
    & starting learning rate & $10^{-2}$ & $10^{-2}$ & $10^{-3}$ & $10^{-4}$ \\
    \cline{2-6}
    & ending learning rate & $10^{-7}$ & $10^{-5}$ & $5 \times 10^{-7}$ & $10^{-7}$ \\
    \cline{2-6}
    & initial PBO's parameters std & $5 \times 10^{-6}$ & $5 \times 10^{-6}$ & $5 \times 10^{-7}$ & $5 \times 10^{-7}$ \\
    \hline
\end{tabular}
\end{table}
\subsubsection{Chain-Walk}\label{A:chain_walk}
We consider all the possible state-action pairs $10$ times as the initial dataset of samples $\mathcal{D}$. These $10$ repetitions help the algorithms to grasp the randomness of the environment. The optimal action-value function $Q^*$ is computed with dynamic programming. The $\ell^2$-norm is computed by taking the norm of the vector representing the action-value function. In Figure \ref{F:chain_walk_v}, we only report the mean value, since the standard deviations over the seeds are negligible. 

\subsubsection{Linear Quadratic Regulator}\label{A:lqr}
The dynamics of the system is $\mathcal{P}(s,a) = -0.46 s + 0.54 a$ and the reward function is $\mathcal{R}(s, a) = -0.73 s^2 - 0.63 sa - 0.93 a^2$. $M$ is chosen to be equal to $-1.20$
The set $\mathcal{D}$ is collected on a mesh of size $11$ by $11$ on the state-action space going from $-4$ to $4$ in both directions, which means that samples belong to $[-4, 4] \times [-4, 4] \subset \sS \times \sA$.
We choose to parameterize the value functions in a quadratic way, thus allowing us to compute the maximum in closed form. However, we assume that the algorithms do not have this knowledge since it is not the case in more general settings. For this reason, we discretize the action space in a set of $200$ actions going from $-8$ to $8$. The architecture of the parameterized PBO trained with ProFQI is a fully connected network composed of one hidden layer having $8$ neurons. Rectified linear unit (ReLU) is used as the activation function. In Figure \ref{F:lqr_w}, we only report the mean value, since the standard deviations over the seeds are negligible.

\subsubsection{Car-On-Hill}\label{A:car_on_hill}
\begin{figure}
    \centering
    \begin{subfigure}{0.469\textwidth}
        \begin{center}
            \includegraphics[scale=.5]{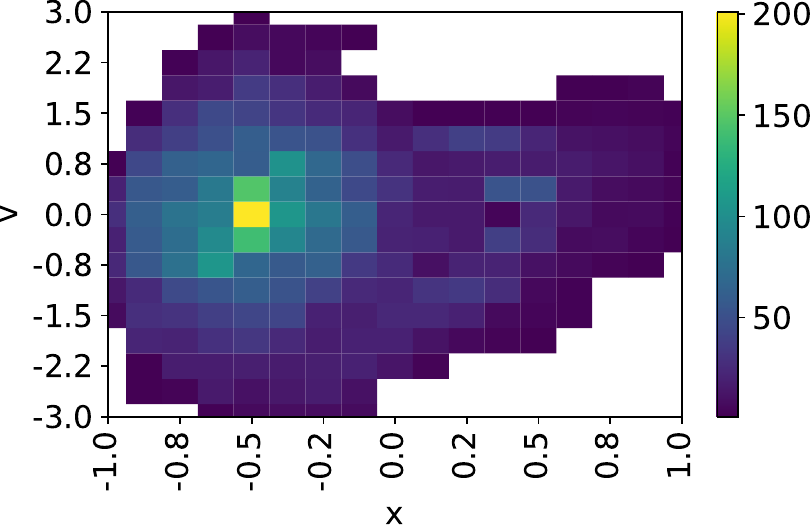}
            \caption{Sample distribution.}
            \label{F:car_on_hill_samples}
        \end{center}
    \end{subfigure}
    \hspace{0.05\textwidth}
    \begin{subfigure}{0.469\textwidth}
        \begin{center}
            \includegraphics[scale=.5]{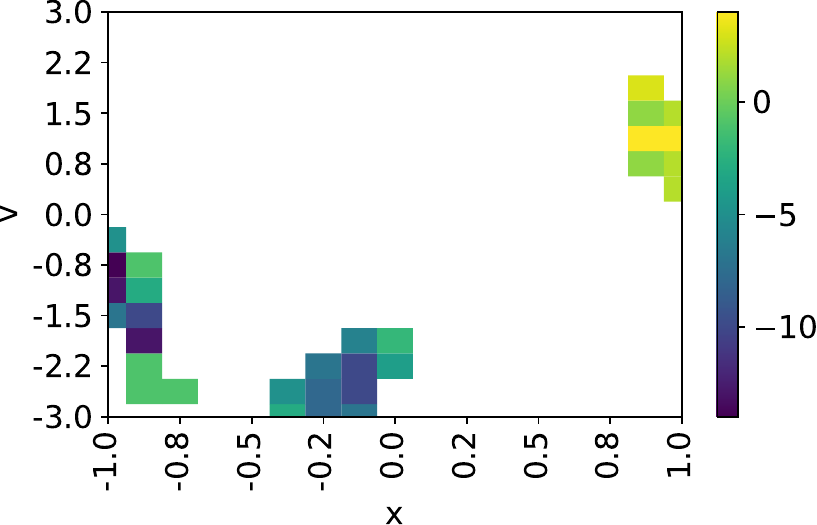}
            \caption{Reward distribution.}
            \label{F:car_on_hill_rewards}
        \end{center}
    \end{subfigure}
    \caption{Composition of the dataset of samples $\mathcal{D}$ on car-on-hill.}
    \label{F:car_on_hill_samples_rewards}
\end{figure}

The agent chooses between $2$ actions: \texttt{left} or \texttt{right}. The state space is $2$-dimensional: position in $[-1, 1]$ and velocity $[-3, 3]$. If the agent succeeds to bring the car up the hill -- at a position greater than $1$ and velocity in between $-3$ and $3$ -- then the reward is $1$, if the agent exceeds the state space, the reward is $-1$; otherwise, the reward is $0$.\\
As our ProFQI is an offline algorithm, we need to make sure that all the necessary exploration has been done in the dataset of samples. For that reason, we first consider a uniform sampling policy to collect episodes starting from the lowest point in the map ($[-0.5, 0]$) with a horizon of $100$. This sampling process is stopped when $4.500$ samples are gathered. To get more samples with positive rewards, we sample new episodes starting from a state located randomly between $[0.1, 1.3]$ and $[0.5, 0.38]$ with a uniform policy as well. In total, $5.500$ samples are collected. The sample and reward distributions over the state space is shown in Figure \ref{F:car_on_hill_samples_rewards}. The action-value functions are parameterized with one hidden layer of $30$ neurons with ReLU as activation functions. The architecture of the parameterized PBO has $4$ hidden layers of $302$ ($2$ times the number of parameters of the action-value functions) neurons each with ReLU as activation functions. To help the training, the value functions are taking actions in $\{-1, 1\}$ instead of the usual $\{0, 1\}$. Given that the policies, the reward, and the dynamics, are deterministic, we perform only one simulation to generate Figure~\ref{F:car_on_hill_J}.

\subsection{Online Experiments}
\begin{table}
\centering
\caption{Summary of all parameters used in the online experiments.}\label{T:parameters_online}
\begin{tabular}{|| c | c || c | c ||} 
    \hline
    \multicolumn{2}{|| c ||}{} & Bicycle & Lunar Lander \\ 
    \hline \hline
    \multicolumn{2}{|| c ||}{horizon} & $50.000$ & $1.000$ \\
    \hline
    \multicolumn{2}{|| c ||}{$\gamma$} & $0.99$ & $0.99$ \\
    \hline
    \multicolumn{2}{|| c ||}{\#$\mathcal{D}$} & $10.000$ & $1.000$ \\
    \hline
    \multicolumn{2}{|| c ||}{$\max$ \#$\mathcal{D}$} & $10.000$ & $20.000$ \\
    \hline
    \multicolumn{2}{|| c ||}{batch size on $\mathcal{D}$} & $500$ & $500$ \\
    \hline
    \multicolumn{2}{|| c ||}{\#steps per update} & $2$ & $2$ \\
    \hline
    \multicolumn{2}{|| c ||}{starting $\epsilon$} & $1$ & $1$ \\
    \hline
    \multicolumn{2}{|| c ||}{ending $\epsilon$} & $10^{-2}$ & $10^{-2}$ \\
    \hline \hline
    \multirow{3}{*}{DQN} 
    & \#fitting steps & $6.000$ & $6.000$ \\
    \cline{2-4}
    & starting learning rate & $10^{-4}$ & $10^{-3}$ \\
    \cline{2-4}
    & ending learning rate & $10^{-6}$ & $10^{-5}$ \\
    \hline  \hline
    \multirow{7}{*}{ProDQN} 
    & \#$\mathcal{W}$ & $30$ & $30$ \\
    \cline{2-4}
    & batch size on $\mathcal{W}$ & $30$ & $15$ \\
    \cline{2-4}
    & \#epochs & $4.000$ & $3.000$ \\
    \cline{2-4}
    & \#training steps & $25$ & $25$ \\
    \cline{2-4}
    & starting learning rate & $10^{-5}$ & $10^{-5}$ \\
    \cline{2-4} 
    & ending learning rate & $10^{-7}$ & $5 \times 10^{-7}$ \\
    \cline{2-4}
    & initial PBO's parameters std & $5 \times 10^{-7}$ & $5 \times 10^{-7}$ \\
    \hline
\end{tabular}
\end{table}
\subsubsection{Bicycle}\label{A:bicycle}
We consider the bicycle problem, as described in~\citet{randlov1998learning}. The state space is composed of $4$ dimensions: $(\omega, \dot{\omega}, \theta, \dot{\theta})$ where $\omega$ is the angle between the floor and the bike, and $\theta$ is the angle between the handlebar and the perpendicular axis to the bike. The goal is to ride a bicycle for $500$ seconds ($50.000$ steps). The agent can apply a torque $T \in \{-2, 0, 2\}$ to the handlebar to make it rotate. The agent can also move its center of gravity in the direction $d \in \{-0.02, 0, 0.02\}$ perpendicular to the bike. As in \citet{lagoudakis2003least}, the agent chooses between applying a torque or moving its center of gravity, resulting in $5$ actions instead of $9$. Usually, a uniform noise in the interval $[-0.02, 0.02]$ is added to $d$. For the purpose of this work, we reduce the number of samples by making the magnitude of the noise $10$ times smaller. A reward of $-1$ is given when the bike falls down, i.e., $|\omega| > 12^\circ$. We use reward shaping to have more informative samples, and we add a reward proportional to the change in $\omega$, i.e., $10^{4} (|\omega_t| - |\omega_{t + 1}|)$, as in~\citet{lagoudakis2003least}. The dataset of samples is composed of $3.500$ episodes starting from a position close to $(0, 0, 0, 0)$ and cut after $20$ steps~\citep{lagoudakis2003least}. The action-value functions are parameterized with one hidden layer of $30$ neurons with ReLU activations. The architecture of the parameterized PBO is composed of $3$ hidden layers of $302$ neurons ($2$ times the number of parameters of the action-value functions) each with ReLU functions as activation functions. During sampling, we only leave the algorithms $20$ steps to explore before ending the episode like in the offline setting. This is done to avoid exploring regions in the state space that are not useful for solving the environment. $100$ simulations are done for each seed shown in Figure \ref{F:bicycle_online_J}, all of them starting from the state $(0, 0, 0, 0)$, i.e., the bicycle standing straight.

\subsubsection{Lunar Lander}
Lunar lander, introduced in \cite{brockman2016openai}, is an environment in which the goal is to make a lunar module land at a specific location while behaving safely for the crew and the rocket. The state space is composed of $8$ dimensions: the position of the rocket, the linear velocities, the angle with the horizon, the angular velocity, and two booleans for each leg being activated when they touch the ground. The action space consists of $4$ actions: fire the main engine, fire the left engine, fire the right engine, and do nothing. The action-value functions are parameterized with $2$ hidden layers of $30$ neurons each with ReLU as activation functions. The architecture of the parameterized PBO is composed of $4$ hidden layers of $2522$ neurons ($2$ times the number of parameters of the action-value functions) each with ReLU functions as activation functions. $100$ simulations are done for each seed shown in Figure \ref{F:bicycle_online_J} each of them starting from a random initial position.

\end{document}

%% file: math_commands.tex
\usepackage{amsmath,amsfonts,bm}

\def\eqref#1{equation~\ref{#1}}

\def\1{\bm{1}}

\def\vmu{{\bm{\mu}}}
\def\vphi{{\bm{\phi}}}
\def\vtheta{{\bm{\theta}}}
\def\vomega{{\bm{\omega}}}
\def\vsigma{{\bm{\sigma}}}

\DeclareMathAlphabet{\mathsfit}{\encodingdefault}{\sfdefault}{m}{sl}
\SetMathAlphabet{\mathsfit}{bold}{\encodingdefault}{\sfdefault}{bx}{n}

\def\sA{{\mathcal{A}}}

\def\sN{{\mathbb{N}}}

\def\sS{{\mathcal{S}}}

\newcommand{\E}{\mathbb{E}}

\newcommand{\R}{\mathbb{R}}

\DeclareMathOperator*{\argmax}{arg\,max}
\DeclareMathOperator*{\argmin}{arg\,min}